%% file: main.tex
\documentclass{article}

\usepackage[table]{xcolor}
\usepackage[final]{corl_2026} 
\usepackage{multirow}
\usepackage{marvosym}
\usepackage{booktabs}
\usepackage{enumitem}
\usepackage{amsmath}
\usepackage{amssymb}
\usepackage{mathtools}
\usepackage{amsthm}
\usepackage{booktabs}
\usepackage{tabularx}
\usepackage{wasysym}

\title{LaST-HD: Learning Latent Physical Reasoning from Scalable Human Data for Robot Manipulation}

\author{
Jiaming Liu$^{*}$$^{\dagger}$\textsuperscript{\rm 1} 
Yinxi Wang\thanks{Equal contribution, $^{\dagger}$Project lead, \textsuperscript{\Letter}Corresponding author.} \hspace{0.1mm} \textsuperscript{\rm 1} 
Chenyang Gu$^{*}$\textsuperscript{\rm 1}
Siyuan Qian$^{*}$\textsuperscript{\rm 1} 
Xiangju Mi$^{*}$\textsuperscript{\rm 1} 
\textbf{Hao Chen$^{\dagger}$\textsuperscript{\rm 2}} \\
\textbf{Jiawei Chen\textsuperscript{\rm 1}} 
\textbf{Qingpo Wuwu\textsuperscript{\rm 1}} 
\textbf{Xiaoqi Li\textsuperscript{\rm 1}} 
\textbf{Nuowei Han\textsuperscript{\rm 1}} 
\textbf{Yiming Zhang\textsuperscript{\rm 1}} 
\textbf{Xuheng Zhang\textsuperscript{\rm 1}} \\
\textbf{Yang Yue\textsuperscript{\rm 4}} 
\textbf{Yeqing Yang\textsuperscript{\rm 4}} 
\textbf{Lei Wang\textsuperscript{\rm 3}} 
\textbf{Peng Jia\textsuperscript{\rm 3}} 
\textbf{Hao Tang\textsuperscript{\rm 1}} 
\textbf{Shanghang Zhang\textsuperscript{\rm 1}\textsuperscript{\Letter}}
\vspace{0.2cm}\\
\textsuperscript{\rm 1} State Key Laboratory of Multimedia Information Processing, School of Computer Science, \\Peking University
\textsuperscript{\rm 2} The Chinese University of Hong Kong\\
\textsuperscript{\rm 3} Simplexity Robotics 
\textsuperscript{\rm 4} Aether Tech
\vspace{0.2cm}\\
Project page: \url{https://siriyep.github.io/last-hd-project-page/}
}

\begin{document}
\maketitle


\begin{abstract}
\input{section/abstract}

\end{abstract}



\input{section/introduction}

\input{section/relatedwork}

\input{section/method}

\input{section/experiment}

\input{section/conclusion}




\bibliography{example}  

\input{section/appendix}

\end{document}

%% file: section/abstract.tex
Human-hand demonstrations provide a direct and scalable source of physical interaction data for robot learning. While manual retargeting is indispensable for establishing kinematic action correspondence across different morphologies, robust transfer requires going beyond geometry to address the underlying alignment of physical dynamics between human and robot manipulation. To address this, we introduce LaST-HD, a novel human-to-robot action learning paradigm that extends reasoning-before-acting VLA by aligning human-hand and robot demonstrations in a shared latent reasoning space. Rather than mimicking human kinematics, LaST-HD trains an auxiliary action-conditioned world model on unpaired human-hand and robot trajectories to synthesize unified latent targets. After aligning cross-embodiment representations in this shared forward-dynamics space, these targets supervise LaST-HD’s latent reasoning process, enabling it to internalize shared physical dynamics and drive efficient human-hand action learning. Moreover, we develop Out-of-Lab (OOL) Glove, a low-cost motion-capture glove tailored to LaST-HD for human-hand data collection. The captured human data provide precise keypoints and serve as universal action supervision across grippers and dexterous hands. Armed with the aligned latent space and high-fidelity human-hand data, we develop a progressive mixed-to-human training recipe comprising mixed human-robot co-training and human-hand online correction post-training. Through mixed co-training, LaST-HD improves generalization to novel objects, scenes, and positions using only human-hand demonstrations. With online correction, LaST-HD further adapts to novel environments and achieves over 90\% accuracy using only 20 minutes of OOL glove data.

%% file: section/introduction.tex
\section{Introduction} 
\label{sec:introduction}

\begin{figure*}[t]
    \centering
    \vspace{-0.2cm}
    \includegraphics[width=0.99\textwidth]{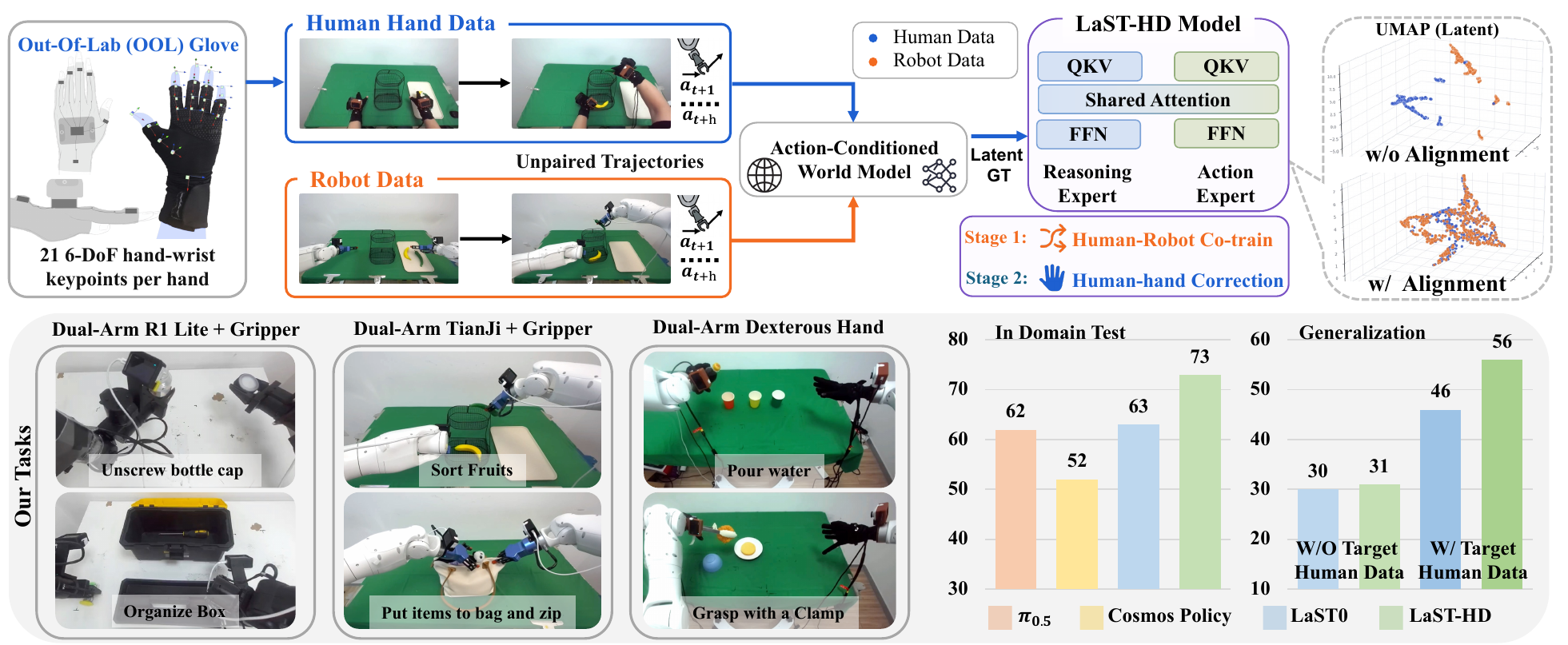} 
    \vspace{-0.2cm}
    \caption{\textbf{Overview of LaST-HD.} 
    LaST-HD aligns human-hand and robot demonstrations through action-conditioned latent physical reasoning, enabling morphology-agnostic action learning under a progressive mixed-to-human training recipe. To support high-quality human data collection, we introduce an OOL glove tailored to LaST-HD. We evaluate LaST-HD on six real-world tasks spanning three embodiments, including dual-arm grippers and dexterous hands.
    }
    \label{fig:teaser}
    \vspace{-0.6cm}
\end{figure*}

Developing generalist manipulation policies in unstructured environments remains a grand challenge. Driven by large-scale pretrained Vision-Language Models (VLMs)~\citep{karamcheti2024prismatic, bai2025qwen3vltechnicalreport, beyer2024paligemma}, Vision-Language-Action (VLA) models~\citep{kim2024openvla, black2024pi_0, liu2025hybridvla, chen2025fast} have emerged as a powerful paradigm for robotic control.
However, scaling these models requires massive robot demonstration datasets~\citep{khazatsky2024droid, wu2024robomind, open_x_embodiment_rt_x_2023}, which are notoriously inefficient to collect due to the labor-intensive nature and substantial hardware overhead of real-robot teleoperation. In contrast, human hand demonstrations offer a direct, data-rich alternative with diverse physical interaction priors, providing an ideal substrate for transferring human manipulation experience to robot manipulation~\citep{chen2021learning,bahl2022human}.

Existing human-to-robot transfer methods have predominantly centered on kinematic retargeting~\citep{qiu2025humanoid, guzey2025bridging} or morphological translation~\citep{lepert2025phantom}, mapping human hand poses directly onto robotic joint spaces. Moving beyond such rigid mappings, traditional policies typically extract visual representations~\citep{kareer2025egomimic, nair2022r3m} or object-centric trajectory priors~\citep{bharadhwaj2024track2act, bahl2022human} from human videos to guide robot learning. In the VLA field, effectively leveraging large-scale human hand data has become even more critical to unlocking open-world generalization~\citep{generalist2026gen1,intelligence2026pi}. 
Recent VLA methods have begun to treat the human hand as a distinct cross-embodiment variant and rely on joint co-training to learn shared representations~\citep{kareeremergence,from_human_skill_to_robotic_mastery_2026,bi2026h,zheng2026egoscale}.
Nevertheless, these co-training approaches either remain sensitive to data scale or overlook the alignment of physical dynamics between human and robot manipulation. 
This raises a question: Can we use the physical reasoning of VLA models as an intermediate interface to better transfer human-hand demonstrations into robot action learning?

In this paper, as shown in Figure~\ref{fig:teaser}, we present LaST-HD, a novel \textbf{human-to-robot action learning paradigm} that extends ``reasoning-before-acting" VLA by using aligned latent reasoning as an intermediate interface for morphology-agnostic action learning.
Rather than relying solely on action-level co-training, LaST-HD embeds glove-collected human hand and robot trajectories into a shared compact latent space to model their underlying physical dynamics.
Building on a Mixture-of-Transformers (MoT) VLA model~\citep{liu2026last}, LaST-HD performs physical reasoning in an aligned latent space and uses shared attention to guide action execution.
To supervise this shared latent space, we employ an action-conditioned world model~\citep{guo2025ctrl} trained on unpaired human and robot trajectories.
Using action labels as weak anchors, the world model aligns human and robot latent features through predicted physical consequences.
Rather than relying on future-frame visual representations, we extract forward-dynamics features from the world model as explicit latent targets for LaST-HD's reasoning expert.
This simple yet deliberate design enables our LaST-HD VLA model to reason about morphology-agnostic physical dynamics, spearheading efficient action learning from human hand data.
In Figure~\ref{fig:teaser}, the UMAP visualizations~\citep{mcinnes2018umap} show that LaST-HD forms structured alignments between human-hand and robot trajectories from the same task.

To scale human interaction data, we develop \textbf{Out-of-Lab (OOL) Glove}, tailored to LaST-HD for low-cost and high-fidelity human-hand data collection.
Unlike conventional exoskeleton gloves that mainly record device-specific joint angles, OOL Glove is designed to capture native human-hand interactions with the environment.
Specifically, it uses IMU-based 6-DoF keypoint tracking to localize 21 hand-wrist keypoints with a sub-millimeter average per-keypoint RMS position error.
The collected data can be retargeted from parallel grippers to dexterous hands, providing teleoperation-level supervision for robot action learning.
Armed with LaST-HD’s latent alignment method and high-quality human-hand data, we further propose a progressive \textbf{mixed-to-human training recipe}.
Specifically, we first co-train LaST-HD on a mixed human-robot corpus, where aligned physical reasoning drives cross-embodiment action generation.
As shown in Figure~\ref{fig:teaser}, our proposed system expands the frontier of human action learning, significantly improving generalization to novel objects, scenes, and positions using only human-hand demonstrations, without requiring target-domain robot data.
After co-training, we introduce the first human-hand online correction strategy, which injects targeted human corrections at failure-prone states. By replacing cumbersome teleoperation with lightweight human-hand corrections, this post-training stage enables rapid adaptation and drives the LaST-HD policy to over 90\% accuracy in novel environments with only 20 minutes of data. 
\textbf{Our overall system contributions are summarized as follows:}

\begin{itemize}[nosep, leftmargin=*]

\item We propose LaST-HD, a novel human-to-robot action learning paradigm that extends reasoning-before-acting VLA by aligning human-hand and robot data in a shared latent physical reasoning space, thereby facilitating morphology-agnostic action learning.

\item We develop OOL Glove, a low-cost wearable platform tailored to LaST-HD for capturing high-quality human-hand interactions as retargetable supervision for both grippers and dexterous hands.

\item We develop a mixed-to-human training recipe that activates LaST-HD's full potential by combining mixed human-robot co-training with human-hand online correction for rapid adaptation.

\end{itemize}

%% file: section/relatedwork.tex
\section{Related Work} 
\label{sec:related}

\textbf{Vision-Language-Action (VLA) Models.}
Driven by large-scale pretrained Vision-Language Models (VLMs)~\citep{karamcheti2024prismatic, bai2025qwen3vltechnicalreport, beyer2024paligemma}, VLA models have become a leading paradigm for generalizable robotic manipulation. Early systems such as RT-series models and OpenVLA~\citep{zitkovich2023rt, kim2024openvla} demonstrated strong cross-task generalization from large-scale training data. Subsequent work improved action generation through efficient tokenization~\citep{chen2025fast, kim2025fine}, flow matching~\citep{black2024pi_0, bjorck2025gr00t, su2026freqpolicy}, and diffusion-based decoders~\citep{wen2024diffusion, liu2025hybridvla, chen2025fast}. More recently, ``reason-before-act'' VLAs have incorporated textual CoT for task planning~\citep{intelligence2025pi05visionlanguageactionmodelopenworld, lin2025onetwovla, zawalski2025roboticcontrolembodiedchainofthought}, future visual or multimodal prediction~\citep{zhao2025cot, gao2025adaworld, zhang2025dreamvla, liu2025mla, cen2025worldvla, gu2025manualvla}, and latent-space reasoning for physical dynamics modeling~\citep{huang2025thinkact, liu2026last, cai2026internvla, lyu2026lda}.
Meanwhile, scaling VLA training with human hand data has emerged as a promising route toward action generalization~\citep{generalist2026gen1, intelligence2026pi, kareeremergence}.

\textbf{Learning from Human data.}
Human hand demonstrations provide a scalable source of physical interaction~\citep{bahl2022human, chen2021learning, banerjee2025hot3d, grauman2024ego}. Early works mainly used egocentric corpora~\citep{grauman2022ego4d, damen2018scaling} to learn robot visual representations~\citep{nair2022r3m, ma2022vip, vc2023}. Later methods extracted more action-proximal supervision from human videos, including keypoint or point-track prediction~\citep{bharadhwaj2024track2act, wang2023mimicplay}, object affordances~\citep{bahl2022human, bahl2023affordances}, and latent actions~\citep{ye2024latent, gao2026dreamdojo, lyu2026lda}; others collected aligned human-robot action pairs with AR/VR systems~\citep{duan2023ar2d2, park2024dexhub}. 
Recent work treats the human hand as another embodiment: EgoMimic~\citep{kareer2025egomimic} and DexWild~\citep{tao2025dexwild} rely on explicit kinematic alignment for unified imitation learning, EgoVLA~\citep{yang2025egovla} transfers pretrained human-motion VLAs via inverse kinematics, while H-RDT~\citep{bi2026h} and EgoScale~\citep{zheng2026egoscale} bridge the embodiment gap through large-scale cross-embodiment pre-/mid-training.
Nevertheless, these methods typically rely on action- or representation-level alignment, which can remain sensitive to data scale. In contrast, LaST-HD makes the first attempt to align human-hand and robot demonstrations through latent physical reasoning, thereby facilitating downstream action learning.

%% file: section/method.tex
\section{Methodology} 
\label{sec:method}

In this section, we present LaST-HD framework for enhancing human hand action learning, consisting of human-to-robot alignment, the OOL Glove design, and a mixed-to-human training recipe.

\subsection{Preliminaries}
\label{sec:MA}

\textbf{Problem Formulation.}
We formulate robotic manipulation as probabilistic sequential decision making~\citep{kim2024openvla}. Given a language instruction $\mathbf{l}$ and visual observation $\mathbf{I}_t \in \mathbb{R}^{H \times W \times 3}$, the policy $\pi_\theta$ predicts an action chunk $\mathbf{a}_{t+1:t+H} \sim \pi_\theta(\cdot \mid \mathbf{I}_t, \mathbf{l})$. The action space is embodiment-specific: dual-arm grippers use two 7-DoF end-effector actions, each consisting of relative translation, Euler-angle rotation, and a binary gripper command, while dexterous hands additionally include hand joint angles, yielding, for example, a 26-DoF action space when equipped with the 20-joint WUJI Hand.

\textbf{Model Architecture.}
As shown in Figure~\ref{fig:method}(a), LaST-HD adopts a Mixture-of-Transformers (MoT) VLA architecture built upon Janus-Pro~\citep{chen2025janus}. Following a ``reasoning-before-acting'' paradigm~\citep{huang2025thinkact, liu2026last}, it first produces compact latent reasoning to model task-relevant physical dynamics before predicting robot actions.
For vision encoder, each observation $\mathbf{I}_{t} \in \mathbb{R}^{H \times W \times 3}$ (H = W = 384), we use SigLIP-Large~\citep{zhai2023siglip} to extract visual features
$f_{\text{img}} \in \mathbb{R}^{N_{\text{img}} \times d_v}$,
where $N_{\text{img}}$ is the number of visual tokens, and $d_v$ is the visual embedding dimension.
These features are projected into the large language model (LLM) hidden space via an MLP.
For the VLA backbone, we adopt DeepSeek-LLM 1.5B and repurpose its 24-layer decoder-only transformer into an MoT-based policy with two distinct components: a reasoning expert and an action expert.
The reasoning expert autoregressively predicts a sequence of latent states $\mathcal{Z} \in \mathbb{R}^{N_{\text{lat}} \times d_l}$, while the action expert predicts the action chunk $\mathbf{a}_{t:t+H-1}$ through flow-matching, where $N_{\text{lat}}$ is the number of reasoning tokens and $d_l$ is the LLM hidden dimension. 
Latent reasoning knowledge is transferred from the reasoning expert to the action expert through a shared attention design. This MoT formulation provides an effective interface for injecting morphology-agnostic physical reasoning priors before action prediction.

\subsection{Human-to-Robot Latent Alignment}
\label{sec:HRLA}

To efficiently train LaST-HD, we leverage scalable human-hand demonstrations, yet the substantial embodiment gap between human hands and robot arms makes direct transfer challenging. Instead of naive action-level co-training, which suffers from severe domain mismatch, we introduce a human-to-robot latent alignment strategy that embeds both domains into a shared physical reasoning space, allowing aligned latent reasoning to support efficient action learning.

\begin{figure*}[t]
    \centering
    \vspace{-0.3cm}
    \includegraphics[width=\textwidth]{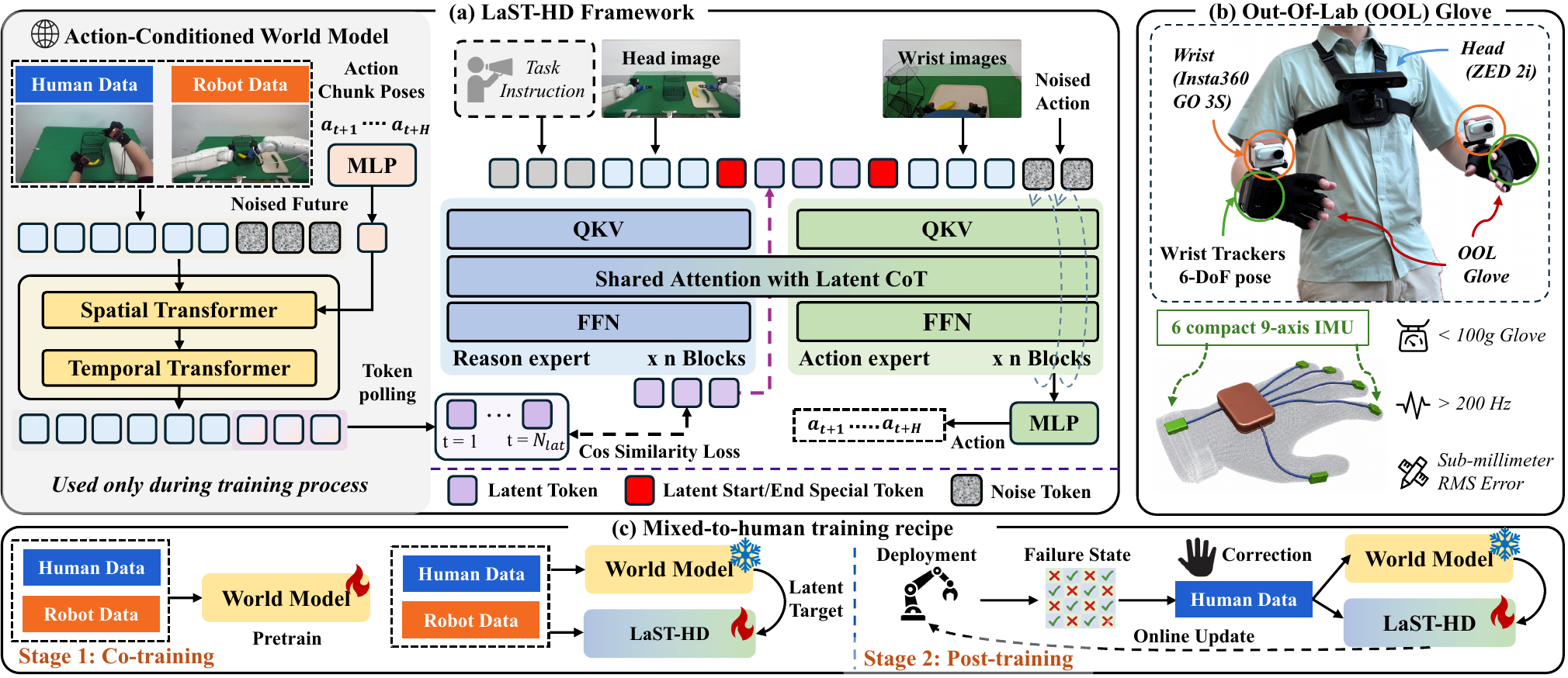} 
    \vspace{-0.7cm}
    \caption{\textbf{Framework.} (a) LaST-HD aligns human-hand and robot data through an action-conditioned world model, which converts predicted physical consequences into denoised future-frame features as latent supervision for the reasoning expert. The aligned latent CoT conditions the action expert through shared attention for action generation.
    (b) We present the OOL Glove and collection platform for human-hand data collection.
    (c) A progressive mixed-to-human training recipe is introduced, where mixed human-robot co-training is followed by human-hand online corrections.
    }
    \label{fig:method}
    \vspace{-0.4cm}
\end{figure*}

\textbf{World Model as Alignment Bridge.} 
To construct this shared latent space, as shown in Figure~\ref{fig:method} (a), we fine-tune an action-conditioned world model~\citep{guo2025ctrl}, on a mixed dataset of glove-collected human and real-robot demonstrations. Importantly, demonstrations from the two domains need not be strictly paired. Specifically, we feed visual observations into the world model, injecting continuous action chunks via cross-attention into every layer to guide generation. At the final denoising step, we extract features from the deepest U-Net layer, which capture predictive physical dynamics while providing domain-invariant representations~\citep{luo2023diffusion,tang2023emergent}. These spatiotemporal features are projected to the LaST-HD latent dimension $d_l$ with an MLP-based aligner, flattened, and compressed by adaptive average pooling into $N_{\text{lat}}$ latent tokens. The resulting tokens serve as explicit ground-truth for supervising LaST-HD’s reasoning expert.
Notably, we use the world model for latent supervision rather than direct action prediction, since its latent features are not sufficiently compact for efficient control, and action conditioning may introduce information leakage.

\textbf{Insights on Latent Alignment.}
The effectiveness of this alignment strategy stems from the physical invariance underlying manipulation tasks. Although human and robot trajectories are not explicitly paired and differ in both kinematic structure and visual appearance, action labels from both domains serve as weak anchors. Unlike future-frame visual representations, which primarily capture appearance evolution, action-conditioned forward-dynamics representations encode the physical consequences of interactions. Since both embodiments obey the same physical laws—for example, pushing an apple produces similar object motion regardless of embodiment—action-conditioned prediction encourages latent alignment around shared task semantics and physical dynamics.
By adopting the “reasoning-before-acting” paradigm, LaST-HD projects human and robot data into a unified latent reasoning space, thereby facilitating action learning from both embodiments.
To validate the effectiveness of this simple yet deliberate alignment design, we conduct UMAP visualizations and attention-map analyses on latent tokens in Appendix~\ref{ap:AQual} and Section~\ref{sec:ablation}, respectively.


\subsection{Human-Hand Data Collection} 
\label{sec:human-hand}

\textbf{Hardware System.}
The Out-of-Lab (OOL) Glove is designed for low-cost, low-latency, and scalable human-hand data collection.
Unlike prior exoskeleton or haptic gloves that impose rigid mechanical constraints, the OOL Glove adopts an ultra-lightweight design, with each glove weighing less than 100 g.
Mechanically, as shown in Figure~\ref{fig:method} (b), the system uses six compact IMU modules to track 20 hand keypoints and one wrist keypoint within a unified hand-centric coordinate frame.
Our glove operates at over 200 Hz with less than 10 ms latency and achieves sub-millimeter average RMS position error per keypoint.
Detailed specifications are provided in Appendix~\ref{ap:ADOG}.

\textbf{Data Collection.}
During data collection, users perform manipulation tasks directly in the physical environment.
As shown in Figure~\ref{fig:method} (b), the overall collection setup includes two wrist-mounted views and one head-/chest-mounted view.
More importantly, OOL Glove provides native human-hand kinematics for LaST-HD training, enabling more efficient learning of physical interactions.
Compared with vision-based bare-hand tracking, this modality provides higher precision; compared with exoskeletons, OOL Glove more directly records human-hand actions, rather than indirectly recording device-specific joint angles. It also enables 4-5 $\times$ faster data collection than robot teleoperation on standard tasks, while avoiding signal loss.
Each collected demonstration forms a synchronized multimodal trajectory consisting of language instructions, observations, and precise hand-wrist states.
After collection, these native hand trajectories are mapped into a unified hand-centric representation.
This representation serves as a general data substrate: gripper commands are derived from fingertip distances, while dexterous-hand joint angles are solved through inverse-kinematics retargeting based on the relative spatial relationships among human-hand keypoints.

\subsection{Mixed-to-Human Training Recipe}
\label{sec: MSTS}

As shown in Figure~\ref{fig:method}(c), we propose a progressive mixed-to-human recipe that fully activates LaST-HD's potential for learning from human-hand data.

\textbf{Stage 1: Mixed Human-Robot Co-training.}
We first train an action-conditioned world model on mixed human-hand and robot trajectories.
As described in Section~\ref{sec:HRLA}, the world-model features corresponding to the predicted future frames serve as morphology-agnostic supervision for LaST-HD.
We find that the world model need not be retrained for every downstream task; including target-embodiment data during pretraining is sufficient to provide aligned latent targets.
For example, our pretraining mixture includes OOL Glove data and Tianji dual-arm data.
During co-training, LaST-HD VLA model is optimized on both human-hand and robot trajectories.
The action expert learns to generate executable robot actions using a flow matching action loss~\citep{black2024pi_0,liu2026last} $\mathcal{L}_{\mathrm{act}}$.
The latent reasoning expert is supervised by the aligned latent targets extracted from the trained world model.
Specifically, given the predicted latent tokens $\hat{\mathbf{z}}_t$ and the latent target $\mathbf{z}^{\mathrm{GT}}_t$, we use a cosine similarity loss:
$
\mathcal{L}_{\mathrm{latent}}
=
\sum_{t=1}^{N_{\mathrm{lat}}}
\left(
1 -
\frac{
\hat{\mathbf{z}}_t \cdot \mathbf{z}_t^{\mathrm{GT}}
}{
\lVert \hat{\mathbf{z}}_t \rVert
\lVert \mathbf{z}_t^{\mathrm{GT}} \rVert
}
\right).
$
The overall LaST-HD model training objective is:
$
\mathcal{L}_{\mathrm{loss}}
=
\mathcal{L}_{\mathrm{act}}
+
\lambda
\mathcal{L}_{\mathrm{latent}},
$
where $\lambda$ balances action and latent supervision. We apply this objective to both pretraining and downstream fine-tuning. Details of the datasets used, including real-world robot pretraining data and OOL glove data, are provided in Appendix~\ref{ap:data}.

\textbf{Stage 2: Human-Hand Online Correction Post-training.}
After mixed co-training, we deploy LaST-HD on the real robot and perform policy rollouts to identify failure-prone states.
Unlike previous post-training methods that collect additional robot teleoperation data~\citep{xu2026compliant}, we collect human-hand corrective demonstrations at failure states using the OOL Glove.
In this stage, the world model is kept frozen.
To incorporate new corrective knowledge while avoiding catastrophic forgetting, we post-train LaST-HD for only $1$--$2$ epochs with balanced replay.
Each batch ($\mathcal{B}$) is constructed by equally sampling from the previous data buffer $\mathcal{D}_{\mathrm{prev}}$ and the human-hand DAgger buffer $\mathcal{D}_{\mathrm{dagger}}$:
$
\mathcal{B}
=
\mathcal{B}_{\mathrm{prev}}
\cup
\mathcal{B}_{\mathrm{dagger}}, 
|\mathcal{B}_{\mathrm{prev}}|
=
|\mathcal{B}_{\mathrm{dagger}}|.
$
The same LaST-HD objective is used for post-training.

%% file: section/experiment.tex
\section{Experiments} 
\label{sec:exp}

Section~\ref{sec:exp-setup} details the experimental setup. Section~\ref{sec:RA} presents comprehensive evaluations of LaST-HD, covering in-domain training scenarios and generalization scenarios. Finally, Section~\ref{sec:ablation} provides ablation studies on key components of our design.

\subsection{Experiment Setup}
\label{sec:exp-setup}

\textbf{Data Collection.}
\label{sec:data collection}
We evaluate LaST-HD on six real-world tasks across three embodiments. For dual-arm parallel-gripper setups, we evaluate (1) \textit{Unscrew Bottle Cap} and (2) \textit{Organize Box} on the Galaxea R1 Lite, alongside (3) \textit{Sort Fruits} and (4) \textit{Put Items to Bag and Zip} on the Tianji Marvin. For dexterous manipulation, we evaluate (5) \textit{Pour Water} and (6) \textit{Grasp with a Clamp} on the Marvin arm equipped with a WUJI hand. All setups utilize three $384 \times 384$ camera views: one head view (ZED 2i) and two wrist views (Insta360 GO 3S).
For each task, we collect 100 in-domain robot teleoperation demonstrations and 50 OOL Glove demonstrations.
We define three generalization scenarios, including unseen positions, objects, and scenes, and collect only 60 OOL Glove demonstrations per scenario for each task.
More details of the hardware setup and data collection are provided in Appendix~\ref{ap:real_world_setup} and Appendix~\ref{ap:data}, respectively.

\textbf{Baselines.}
\label{sec:baselines}
We compare LaST-HD with three representative SOTA methods: LaST$_0$~\citep{liu2026last}, a latent-CoT VLA model; $\pi_{0.5}$~\citep{black2024pi_0}, a strong VLA policy; and Cosmos-Policy~\citep{kim2026cosmos}, a world-action model.
All baselines use official full fine-tuning setups; implementation details are provided in Appendix~\ref{ap:BID}.

\input{tables/experiments_in_all}

\subsection{Result Analysis}
\label{sec:RA}

\textbf{In-Domain Setting.}
\label{sec:indomain}
We evaluate all methods in training-domain scenarios.
All baselines are trained on 100 real-robot demonstrations.
For LaST-HD, we evaluate two variants: LaST-HD trained with 100 robot demonstrations, and LaST-HD (Mix-HD) trained with 50 robot and 50 OOL Glove demonstrations collected under the same in-domain scenario.
Each method is evaluated with 20 rollouts per task.
We report task completion success rates, with success criteria detailed in Appendix~\ref{ap:data}.

Table~\ref{tab:in_domain_all} reports the in-domain results, where both LaST-HD variants achieve the highest average success rates across six complex tasks.
While baselines struggle on multi-step tasks such as \textit{Sort Fruits} and \textit{Put Items into Bag and Zip}, LaST-HD achieves 95\% and 80\% success rates, respectively, benefiting from stronger physical latent reasoning.
The gap further widens in high-DoF dexterous manipulation with the Tianji Marvin + WUJI setup.
LaST-HD (Mix-HD) maintains performance comparable to LaST-HD on four out of six tasks, indicating that OOL Glove demonstrations provide effective supervision for action learning.
These results validate that LaST-HD supports complex physical reasoning across challenging tasks, while mixed-data co-training effectively injects human priors without degrading native robot capabilities.

\input{tables/experiments_out_allall}

\begin{figure*}[!t]
\vspace{-0.2cm}
\includegraphics[width=\textwidth]{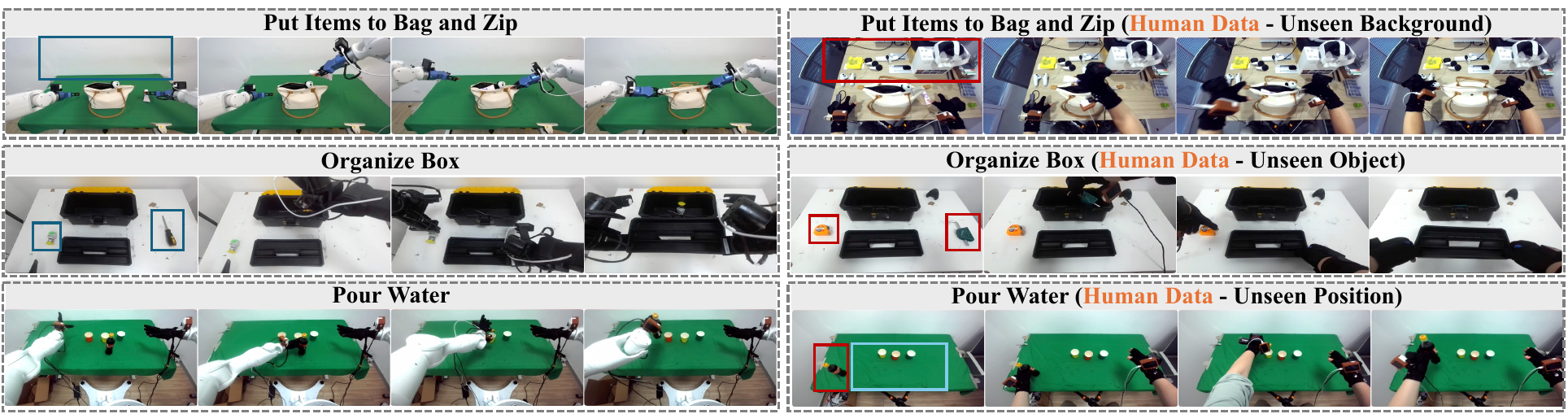}
\centering
\vspace{-0.6cm}
\caption{Robot executions are shown on the left, and OOL Glove data in unseen scenarios on the right.
Red boxes mark unseen configurations, while blue boxes mark in-domain configurations.}
\label{fig:vis}
\vspace{-0.5cm}
\end{figure*}

\textbf{Generalization Setting.}
\label{sec:gen}
To evaluate the generalization capacities, we assess the models under two distinct paradigms.
First, in zero-shot evaluation, $\pi_{0.5}$, Cosmos-Policy, LaST$_0$, and LaST-HD (Mix-HD) are directly tested using their in-domain checkpoints without further fine-tuning.
Second, to assess whether human-hand data alone can enable generalization to unseen scenarios, we compare LaST$_0$ (w/ unseen HD) and LaST-HD (w/ unseen HD), both trained on 100 in-domain robot demonstrations plus 60 human-hand demonstrations per unseen scenario.
We visualize the real-robot evaluation process and unseen scenarios in Figure~\ref{fig:vis}.

As shown in Table~\ref{tab:ood_all}, LaST-HD (w/ unseen HD) achieves substantial improvements over all baselines, showing that aligning human and robot data in a shared latent space enables low-cost human demonstrations to drive robot generalization to new scenarios.
\textbf{Unseen Position:} In the zero-shot setting, all methods show a substantial drop in average success rate, e.g., $\pi_{0.5}$ drops to 12\%.
With target-domain human-hand data, LaST-HD (w/ unseen HD) reaches 41\% average success, demonstrating that human data helps expand the policy's physical exploration space.
\textbf{Unseen Object:} LaST-HD (w/ unseen HD) improves the average success rate to 58\%, outperforming the previous SOTA $\pi_{0.5}$ by 22\% and showing that physical latent reasoning supports robust semantic understanding and adaptive contact patterns.
\textbf{Unseen Background:} LaST-HD (w/ unseen HD) achieves a strong 68\% average success across six tasks, showing that low-cost human-hand data, together with our proposed training recipe, improves robustness to novel visual contexts.

\textbf{Human-Hand Online Correction.}
As shown in Fig.~\ref{fig:ablation}(a), we evaluate human-hand post-training on \textit{Sort Fruits} incrementally: after collecting 10, 20, and 60 OOL Glove trajectories (n), we update LaST-HD and test the resulting policy at each stage.
Collecting 60 OOL Glove trajectories takes only 20 minutes.
With 20 human-hand trajectories, the success rate reaches 100\% for Unseen Background, while with 60 human-hand trajectories, it reaches 100\% for Unseen Object.
Even under the hardest spatial shift, \textit{Unseen Position}, performance improves monotonically from 60\% to 80\%.
These results show that our method, together with OOL Glove data collection, enables rapid adaptation when robots are deployed in new scenarios, while LaST-HD's human-to-robot latent alignment improves the efficiency of learning from human demonstrations.

\subsection{Ablation Study}
\label{sec:ablation}
To validate the key design choices of LaST-HD, we conduct ablations on the dual-arm Sort Fruits task.
All results are averaged success rates evaluated across three generalization settings.

\begin{figure*}[!t]
\vspace{-0.2cm}
\includegraphics[width=\textwidth]{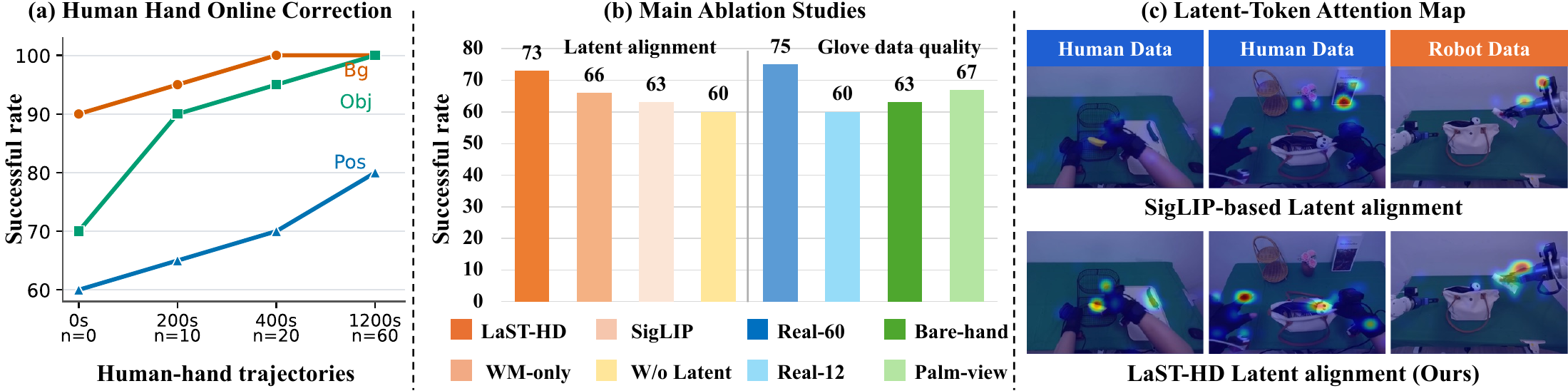}
\centering
\vspace{-0.7cm}
\caption{
\textbf{(a)} Success rate under online correction with varying amounts of correction data.
\textbf{(b)} Main ablation studies on LaST-HD design choices.
\textbf{(c)} Visualization of latent-token attention maps.
}
\label{fig:ablation}
\vspace{-0.25cm}
\end{figure*}

\textbf{Effect of latent alignment strategies.}
We first examine how latent target design affects LaST-HD's mixed human-robot learning.
As shown in the left side of Figure~\ref{fig:ablation}(b), we compare LaST-HD with three variants: WM-only uses world-model latent targets without action conditioning; SigLIP follows LaST$_0$~\citep{liu2026last} and uses future SigLIP image features as latent targets; and W/o Latent removes latent reasoning and retains only the action expert.
For results, removing latent reasoning causes a clear performance drop from 73\% to 60\%, showing that action-level co-training alone cannot fully exploit human-hand demonstrations.
Compared with the SigLIP-based and WM-only variants, using the action-conditioned world-model latents as supervision enables LaST-HD to achieve the best performance.
This indicates that world-model targets capture temporal cues, while action conditioning further anchors human-hand and robot trajectories through predicted physical consequences.
As shown in Fig.~\ref{fig:ablation}(c), LaST-HD shows stronger latent attention on embodiment-object interactions than the SigLIP-based baseline for both human-hand and robot data, indicating more task-relevant and embodiment-agnostic physical reasoning.

\textbf{Effectiveness of OOL Glove data.}
We further evaluate the quality of OOL Glove data against other data collection paradigms under the same LaST-HD training recipe.
As shown in the right part of Fig.~\ref{fig:ablation}(b), across three generalization scenarios, we compare 60 OOL Glove demonstrations with 60 vision-based Bare-hand demonstrations~\citep{zhang2025hawor}, as well as real-robot data using either 12 trajectories per scenario (Real-12), which matches the collection time of 60 human-hand demonstrations, or 60 trajectories per scenario (Real-60).
Implementation details are provided in Appendix~\ref{ap:BID}.
OOL Glove data achieves 73\%, consistently outperforming bare-hand demonstrations (63\%), indicating higher-quality interaction data.
Under the same collection time, our method outperforms real-robot data (Real-12) by 13\%, while remaining comparable to the Real-60 setting.
This shows that our latent alignment strategy enables efficient transfer of human interaction data into robot action learning.
We also find that placing the wrist camera near the thumb-index web space outperforms the palm-view setup (67\%).
Additional ablations on LaST-HD hyperparameters are provided in Appendix~\ref{ap:AQuan}.

%% file: tables/experiments_in_all.tex
\begin{table}[t]
  \centering
  \small
  \vspace{-0.4cm}
  \caption{\textbf{In-domain Results.} We compare LaST-HD against baselines across three distinct robot embodiments with six tasks. (Mix-HD) denote the model trained on a robot-human data mix.}
  \vspace{-0.2cm}
  \label{tab:in_domain_all}

  \setlength{\aboverulesep}{0pt}
  \setlength{\belowrulesep}{0pt}
  \renewcommand{\arraystretch}{1.2}

  \resizebox{\textwidth}{!}{
  \begin{tabular}{l c c c c c c c}
    \toprule
    \multirow{2}{*}{\textbf{Method}} & \multicolumn{2}{c}{\textbf{Galaxea R1 Lite}} & \multicolumn{2}{c}{\textbf{Tianji Marvin}} & \multicolumn{2}{c}{\textbf{Tianji Marvin + WUJI}} & \multirow{2}{*}{\textbf{Avg}} \\
    \cmidrule(lr){2-3} \cmidrule(lr){4-5} \cmidrule(lr){6-7}
    & \textbf{Unscrew Cap} & \textbf{Organize Box} & \textbf{Sort Fruits} & \textbf{Put and Zip} & \textbf{Pour Water} & \textbf{Grasp Clamp} & \\
    \midrule
    $\pi_{0.5}$       & 0.70 & \textbf{0.70} & 0.85 & 0.75 & 0.30 & 0.40 & 0.62 \\
    Cosmos-Policy     & 0.75 & 0.50 & 0.85 & 0.60 & 0.20 & 0.20 & 0.52 \\
    LaST$_0$            & 0.80 & \textbf{0.70} & 0.75 & 0.60 & 0.40 & \textbf{0.50} & 0.63 \\
    \rowcolor{gray!15} \textbf{LaST-HD} & \textbf{0.85} & \textbf{0.70} & \textbf{0.95} & \textbf{0.80} & \textbf{0.60} & 0.45 & \textbf{0.73} \\
    \rowcolor{gray!15} \textbf{LaST-HD (Mix-HD)} & \textbf{0.85} & \textbf{0.70} & 0.85 & \textbf{0.80} & 0.40 & 0.45 & 0.68 \\
    \bottomrule
  \end{tabular}
  }
  \vspace{-0.3cm}
\end{table}

%% file: tables/experiments_out_allall.tex
\begin{table}[t]
  \centering
  \small
  \vspace{-0.3cm}
  \caption{\textbf{Generalization Results.} Task-level success rates across three generalization scenarios, including zero-shot evaluation and adaptation using only OOL Glove human data (w/ unseen HD).
  }
  \vspace{-0.2cm}
  \label{tab:ood_all}

  \setlength{\aboverulesep}{0pt}
  \setlength{\belowrulesep}{0pt}
  \renewcommand{\arraystretch}{1.15}

  \resizebox{\textwidth}{!}{
  \begin{tabular}{l l c c c c c c c c}
    \toprule
    \multirow{2}{*}{\textbf{Method}} & \multirow{2}{*}{\textbf{Scenario}} & \multicolumn{2}{c}{\textbf{Galaxea R1 Lite}} & \multicolumn{2}{c}{\textbf{Tianji Marvin}} & \multicolumn{2}{c}{\textbf{Tianji Marvin + WUJI}} & \multirow{2}{*}{\textbf{Avg}} & \multirow{2}{*}{\textbf{Global Avg}} \\
    \cmidrule(lr){3-4} \cmidrule(lr){5-6} \cmidrule(lr){7-8}
    & & \textbf{Unscrew Cap} & \textbf{Organize Box} & \textbf{Sort Fruits} & \textbf{Put and Zip}  & \textbf{Pour Water} & \textbf{Grasp Clamp} & & \\
    
    \midrule
    \rowcolor{gray!15} \multicolumn{10}{c}{\textbf{Trained on in-domain data only}} \\
    \midrule
    \multirow{3}{*}{$\pi_{0.5}$}
      & Position   & 0.10 & 0.10 & 0.15 & 0.15 & 0.10 & 0.10 & 0.12 & \multirow{3}{*}{0.30} \\
      & Object    & 0.40 & 0.40 & 0.50 & \textbf{0.60} & 0.15 & 0.10 & 0.36 & \\
      & Background & 0.45 & 0.40 & 0.75 & 0.60 & 0.20 & 0.20 & 0.43 & \\
    \midrule
    \multirow{3}{*}{Cosmos-Policy}
      & Position   & 0.10 & 0.15 & 0.35 & 0.00 & 0.10  & 0.05  & 0.13 & \multirow{3}{*}{0.26} \\
      & Object    & 0.60 & 0.10 & 0.60 & 0.10 & 0.20  & 0.05  & 0.28 & \\
      & Background & 0.60 & 0.15 & 0.65 & 0.60 & 0.15 &  0.10 & 0.38 & \\
    \midrule
    \multirow{3}{*}{LaST$_0$}
      & Position   & 0.15 & 0.10 & 0.35 & 0.05 & 0.20 & 0.05  & 0.15 & \multirow{3}{*}{0.30} \\
      & Object    & 0.40 & 0.35 & 0.45 & 0.35 & 0.30  &  0.05 & 0.32 & \\
      & Background & 0.55 & 0.35 & 0.60 & 0.50 & 0.35 & 0.20 & 0.43 & \\
    \midrule
    \multirow{3}{*}{\begin{tabular}{@{}c@{}}
    LaST-HD \\
    (Mix-HD)
    \end{tabular}}
      & Position   & 0.20 & 0.15 & 0.25 & 0.10 & 0.10  & 0.10 & 0.15 & \multirow{3}{*}{0.31} \\
      & Object    & 0.45 & 0.40 & 0.40  &  0.35&  0.40 &  0.10& 0.35 & \\
      & Background & 0.55 & 0.45 & 0.50  & 0.40  & 0.40   & 0.30  & 0.43 & \\
      
    \midrule
    \rowcolor{gray!15} \multicolumn{10}{c}{\textbf{Trained with additional human data in unseen scenarios}} \\
    \midrule
    \multirow{3}{*}{\begin{tabular}{@{}c@{}}
    LaST$_0$ \\
    (w/ unseen HD)
    \end{tabular}}
      & Position   & \textbf{0.30} & \textbf{0.40} & 0.50 & 0.35 & 0.20 &  0.20 & 0.33 & \multirow{3}{*}{0.46} \\
      & Object    & 0.65 & 0.50 & 0.65 & 0.50 & 0.35 & 0.35  & 0.49 & \\
      & Background & \textbf{0.65} & 0.60 & 0.70 & 0.60 & 0.45 &  \textbf{0.45} & 0.58 & \\
    \midrule
    \multirow{3}{*}{\begin{tabular}{@{}c@{}}
    LaST-HD \\
    (w/ unseen HD)
    \end{tabular}}
      & Position   & \textbf{0.30} & \textbf{0.40} & \textbf{0.60} & \textbf{0.45} &  \textbf{0.35} & \textbf{0.35}  & \textbf{0.41} & \multirow{3}{*}{\textbf{0.56}} \\
      & Object    & \textbf{0.75} & \textbf{0.60} & \textbf{0.70} & \textbf{0.60} & \textbf{0.45}  &  \textbf{0.40} & \textbf{0.58} & \\
      & Background & \textbf{0.65} & \textbf{0.70} & \textbf{0.90} & \textbf{0.80} & \textbf{0.60} & \textbf{0.45} & \textbf{0.68} & \\
    \bottomrule
  \end{tabular}
  }
  \vspace{-0.3cm}
\end{table}

%% file: section/conclusion.tex
\section{Conclusion and Limitation}
\label{sec:conclusion}

We presented LaST-HD, a reasoning-before-acting VLA framework that learns robot actions from human-hand data through human-to-robot latent alignment.
By supervising latent reasoning with action-conditioned world-model features, LaST-HD captures morphology-agnostic physical dynamics and transfers human interaction knowledge to robot action generation.
To support scalable and high-fidelity human data collection, we introduced OOL Glove, a low-cost wearable system for capturing native human-hand interactions.
We further proposed a mixed-to-human training recipe that progressively improves LaST-HD's dynamics modeling and action generation, achieving over 90\% accuracy in novel scenarios.
One limitation is that latent reasoning is not yet real-time, while our main insight is to use physical reasoning to improve human action learning.
Future work will explore fast-slow system designs from prior work~\citep{chen2025fast} or further compress the latent space.

%% file: section/appendix.tex
\appendix

\section{Additional Details of OOL Glove}
\label{ap:ADOG}

\subsection{Hardware System}
\begin{figure*}[t]
    \centering
    \includegraphics[width=1.0\textwidth]{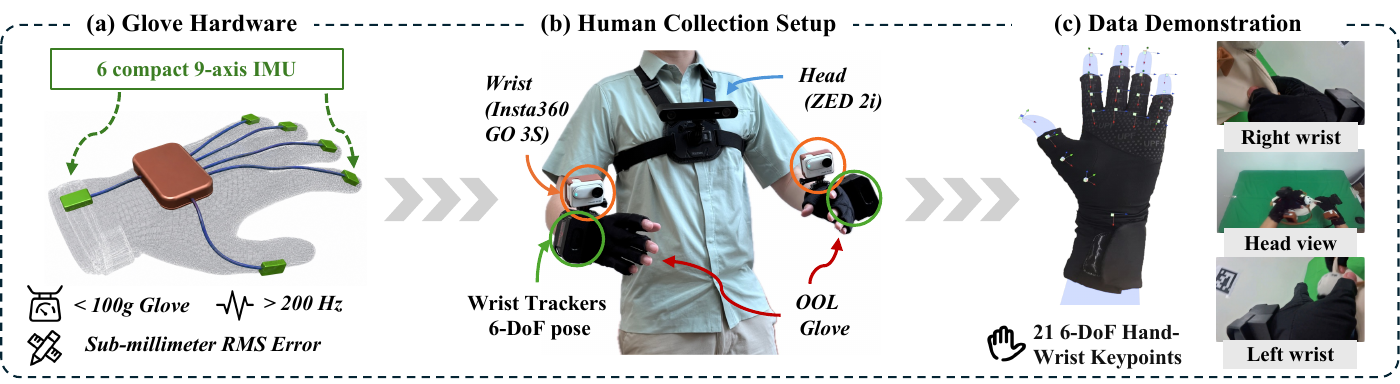} 
    \caption{
    \textbf{OOL Glove hardware and human-native data collection setup. }(a) Lightweight glove hardware with six compact IMU-based sensing modules. (b) Natural manipulation demonstrations are collected with synchronized egocentric vision, wrist 6-DoF tracking, and glove-based hand kinematics. (c) The system reconstructs metric 6-DoF hand-wrist keypoints for retargetable action supervision, operating at high sampling rate and ensuring reliable retargeted robot actions.
    }
    \label{fig:ool_glove_setup}
\end{figure*}

The Out-of-Lab (OOL) Glove is engineered to prioritize scalable and human-native data collection, emphasizing affordability, ease of deployment, and minimal physical interference. The overall hardware and collection setup are illustrated in Fig.~\ref{fig:ool_glove_setup}. While existing exoskeleton or haptic gloves can provide high-degree-of-freedom (DoF) sensing and force feedback for dexterous teleoperation, their rigid linkages, cable transmissions, and active feedback modules inevitably impose additional mechanical constraints on natural hand motion. To address this, the OOL Glove removes bulky mechanical structures in favor of an ultra-lightweight form factor, with each glove weighing less than 100 g, as shown in Fig.~\ref{fig:ool_glove_setup}(a). This design preserves the user's original hand kinematics during real-world object interaction, allowing the collected demonstrations to retain the intrinsic flexibility, compliance, and physical regularities of human manipulation without being filtered through embodiment-specific exoskeleton constraints.

Mechanically, the system utilizes six compact IMU-based 6-DoF sensing modules to accurately capture human hand motion. Following user-specific calibration and rigorous temporal synchronization, the raw IMU sensor streams are mathematically mapped into a unified hand-centric coordinate frame. This framework continuously tracks 20 anatomical hand keypoints alongside one wrist keypoint. As shown in Fig.~\ref{fig:ool_glove_setup}(b) and (c), the human-native collection setup synchronizes egocentric and wrist-view visual observations, wrist 6-DoF poses, and OOL Glove-based hand kinematics during natural hand-object interaction. By converting raw sensor orientations into metric keypoint trajectories, the system faithfully preserves the fine-grained hand-object interaction geometry essential for physical reasoning. 

In terms of performance, the OOL Glove achieves high-fidelity data acquisition suitable for dynamic robotic control. The hardware operates at a sampling rate exceeding 200 Hz with an end-to-end latency of less than 10 ms. Furthermore, the kinematic solver ensures exceptional spatial precision, achieving a sub-millimeter average root-mean-square (RMS) position error per keypoint. Ultimately, these high-quality, metric trajectories provide action-proximal supervision that can be seamlessly retargeted across diverse robot morphologies, spanning from standard parallel grippers to highly articulated dexterous hands.

\subsection{Data Collection}

During data collection, users perform manipulation tasks directly in the physical environment while wearing the OOL Glove. Unlike conventional robot teleoperation systems that require an intermediate human-to-robot control interface (e.g., master arms, space mice, or handheld grippers), OOL Glove directly records native human-hand motions. As a result, the collection process closely matches natural human behavior and substantially improves data collection efficiency. In practice, this allows demonstrations to be collected 4--5$\times$ faster than standard robot teleoperation while preserving rich human interaction dynamics.
Meanwhile, compared with vision-based bare-hand tracking methods, OOL Glove provides more accurate and temporally stable hand-state measurements. Compared with exoskeletons or teleoperation devices, it avoids recording embodiment-specific control signals and instead directly captures human actions.

For wrist-camera placement, the camera is not restricted to a specific location and can be mounted, for example, on the underside of the wrist or the back of the hand. However, we empirically find that positioning the camera near the thumb-index web space provides the most informative viewpoint. This location offers better visibility of object contacts and finger-object interactions, particularly for dexterous manipulation tasks.

Each demonstration yields a synchronized multimodal trajectory consisting of observations, actions, and language instructions. Visual observations and hand-wrist states are recorded directly during data collection. Language instructions can be obtained either by recording spoken task descriptions through a microphone and subsequently refining the transcriptions with a vision-language model (VLM)~\citep{yang2025qwen3}, or by directly annotating demonstrations using a VLM. This process enables scalable acquisition of multimodal robot-learning data with minimal manual effort.

After collection, all demonstrations are converted into a unified hand-centric representation. This representation exhibits strong cross-embodiment reusability. For parallel-jaw grippers, grasp commands can be derived from fingertip distances and wrist trajectories. For dexterous hands, finger motions can be retargeted through kinematic mapping while preserving fine-grained contact geometry and manipulation intent. Consequently, the same human demonstration can be reused across both low-DoF grippers and high-DoF dexterous hands, significantly improving data scalability.
For hand-to-robot retargeting, our current system employs heuristic mappings tailored to each dexterous hand embodiment. Consequently, a new retargeting pipeline must be constructed when introducing a new robotic hand. An important future direction is to replace these manually designed mappings with learnable retargeting models. We also observe that retargeting quality plays a critical role throughout both large-scale pre-training and downstream supervised fine-tuning, where preserving manipulation intent and contact dynamics is essential for successful policy learning.

\section{Real-World Setup}
\label{ap:real_world_setup}

\begin{figure*}[t]
    \centering
    \includegraphics[width=0.9\textwidth]{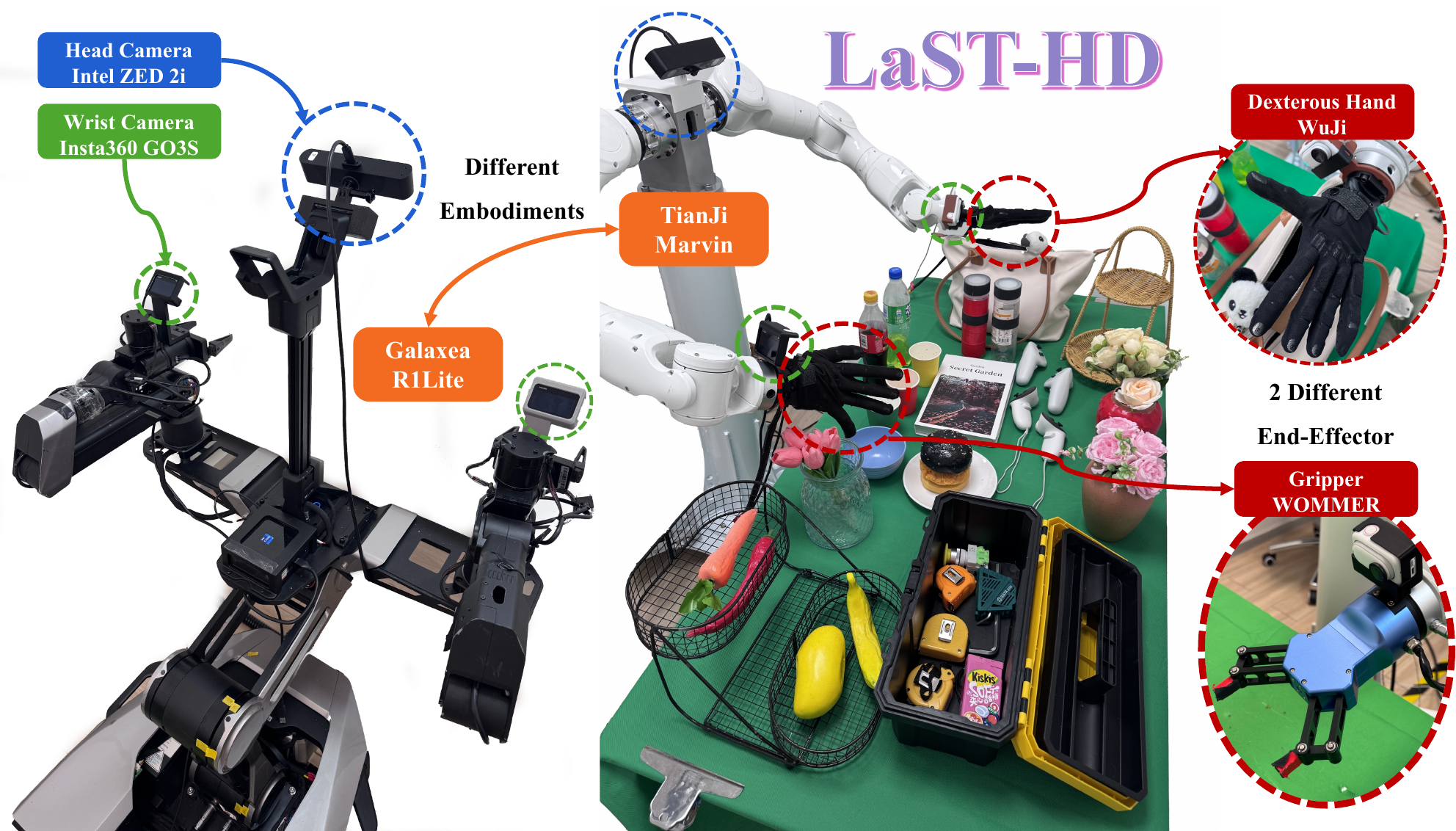} 
    \vspace{-0.1cm}
    \caption{
    \textbf{Real-world setup.}
    Visualization of the two robot embodiments and three actuator configurations used in our real-world experiments, together with the corresponding task assets.
    }
    \label{fig:realset}
    \vspace{-0.2cm}
\end{figure*}

As shown in Figure~\ref{fig:realset}, the physical deployment of LaST-HD spans three real-world robot embodiments, covering both dual-arm parallel-gripper manipulation and high-DoF dexterous control. For the 6-DoF Galaxea R1 Lite platform, we use a dual-arm configuration of right (R) and left (L) arm with parallel grippers (g) for \textit{Unscrew Bottle Cap} and \textit{Organize Box}. The action $\mathbf{a}_m$ is formulated as:
\begin{equation}
    \mathbf{a}_m = [\Delta \theta^R_{1:6}, g^R, \Delta \theta^L_{1:6}, g^L] \in \mathbb{R}^{14}
\end{equation}
For the 7-DoF Tianji Marvin parallel-gripper setup, we evaluate dual-arm manipulation on \textit{Sort Fruits} and \textit{Put Items to Bag and Zip}. The action $\mathbf{a}_m$ is formulated as:
\begin{equation}
    \mathbf{a}_m = [\Delta \theta^R_{1:7}, g^R, \Delta \theta^L_{1:7}, g^L] \in \mathbb{R}^{16}
\end{equation}
For dexterous manipulation, we equip a Tianji Marvin arm with a 20 DoF WUJI dexterous hand (h) and evaluate \textit{Pour Water} and \textit{Grasp with a Clamp}. The action $\mathbf{a}_m$ is formulated as:
\begin{equation}
    \mathbf{a}_m = [\Delta \theta^R_{1:7}, \Delta h^R_{1:20}, \Delta \theta^L_{1:7}, \Delta h^L_{1:20}] \in \mathbb{R}^{54}
\end{equation}
Across all platforms, the observation suite consists of three RGB views resized to $384 \times 384$: one head-mounted ZED 2i camera providing an egocentric scene view, and two wrist-mounted Insta360 GO 3S cameras capturing close-range embodiment-object interactions around the end effectors.

Robot demonstrations are collected through embodiment-specific teleoperation interfaces. For Galaxea R1 Lite, we use its native leader-follower teleoperation system, where a human-operated leader arm with matched kinematics commands the R1 Lite follower arms, enabling intuitive bimanual data collection.
On the Tianji Marvin parallel-gripper setup, operators use SpaceMouse teleoperation to provide 6-DoF Cartesian end-effector commands together with gripper open/close commands. On the Tianji Marvin + WUJI setup, operators wear our OOL Glove: wrist motion drives the Marvin arm, while finger-joint measurements are retargeted online to the WUJI hand, enabling synchronized arm-hand demonstrations. All teleoperated trajectories are synchronized with the three camera streams and stored as language-conditioned observation-action sequences for downstream training and evaluation.

\section{Training Datasets and Evaluation Protocols}
\label{ap:data}
In this section, we describe the large-scale real-robot datasets used for pre-training and the human-hand demonstrations collected with our OOL Glove. We further detail the construction of our six real-world manipulation tasks, including the task setups and success criteria used for evaluation.

\subsection{Large Scale Pre-Training}
\label{ap:pretraining_data}

To endow LaST-HD VLA model with broad motor primitives and physical common sense before downstream human-to-robot co-training~\cite{kim2024openvla}, we adopt a large-scale pre-training mixture consisting of 400K trajectories and 28M frames curated from Open-X-Embodiment~\cite{o2024open}, DROID~\cite{khazatsky2024droid}, and RoboMIND~\cite{wu2024robomind}. Table~\ref{tab:datasets} reports the proportion of each dataset used in the pre-training mixture. Following standard data filtering practices in prior VLA works~\cite{liu2026last}, we remove low-quality trajectories and reconstruct the relative action poses from consecutive robot-state differences to ensure physically consistent action annotations.
Our framework is naturally compatible with both large-scale robot demonstrations and human-hand demonstrations collected by OOL Glove. In principle, both the action-conditioned world model and the LaST-HD VLA model can be pre-trained on the above robot datasets together with our collected OOL Glove human-hand data. As shown in Table~\ref{tab:ooldata}, we have accumulated substantial experience and data resources for OOL Glove collection, covering more than 2,000 hours of demonstrations across diverse real-world scenarios.
However, to ensure fair comparison with prior VLA baselines~\cite{black2024pi_0, liu2026last}, we pre-train the LaST-HD Mixture-of-Transformers (MoT) model only on real-robot trajectories.

To provide explicit supervision for latent physical reasoning, we first pre-train an action-conditioned world model using both OOL Glove human-hand demonstrations and real-robot trajectories. The world model takes visual observations as input and injects the corresponding action chunks through cross-attention, learning to predict future physical consequences across human and robot embodiments. After pre-training, we freeze the world model and use it to precompute latent ground-truth targets for LaST-HD. Specifically, for each trajectory, we extract features from the deepest U-Net layer at the final denoising step, where the features encode action-aware spatiotemporal dynamics. These features are then projected into the LaST-HD latent dimension with an MLP aligner, flattened, and compressed through adaptive average pooling into a fixed number of latent tokens. The impact of the latent-token length is further analyzed through an ablation study in Appendix~\ref{ap:AQuan}. The resulting tokens serve as latent ground-truth targets for supervising the reasoning expert of LaST-HD.

Because these latent targets are computed entirely offline, they are simply loaded together with the standard visual, language, and action inputs during training. This design provides LaST-HD with compact action-conditioned physical reasoning anchors, enabling the model to learn predictive environmental dynamics and establish a shared latent reasoning space across human and robot data. We are continuously collecting in-the-wild OOL Glove demonstrations and plan to release a high-quality human-hand manipulation dataset to support future research on scalable human-to-robot action learning.

\begin{table}[t]
\centering
\small
\renewcommand{\arraystretch}{1.1}
\setlength{\tabcolsep}{6pt} 
\caption{\textbf{Datasets for Pre-training.}
The names of selected datasets for large-scale pretraining and their sampling ratios (\%).}
\vspace{-0.2cm}
\label{tab:datasets}
\begin{tabular}{lc | lc}
    \toprule
    \multicolumn{4}{c}{\textbf{Training Dataset Mixture}} \\
    \midrule
    \textbf{Dataset} & \textbf{Ratio (\%)} & \textbf{Dataset} & \textbf{Ratio (\%)} \\
    \midrule
    BridgeV2~\cite{ebert2022bridge,walke2023bridgedata} & 20.82 & Nyu Franka Play~\cite{cui2022play} & 0.24 \\
    Kuka~\cite{kalashnikov2018qt} & 20.22 & Stanford Hydra~\cite{belkhale2023hydra} & 0.20 \\
    Fractal~\cite{rt12022arxiv} & 13.67 & RoboMIND~\cite{wu2024robomind} & 0.20 \\
    Robo-Net~\cite{dasari2020robonet} & 11.53 & Jaco Play~\cite{dass2023jacoplay} & 0.19 \\
    Language Table~\cite{lynch2023interactive} & 7.72 & Dobb-E~\cite{shafiullah2023dobbe} & 0.18 \\
    BC-Z~\cite{jang2022bc} & 7.54 & Toto~\cite{zhou2023train} & 0.17 \\
    Maniskill~\cite{gu2023maniskill2unifiedbenchmarkgeneralizable} & 5.26 & Furniture Bench~\cite{heo2023furniturebench} & 0.09 \\
    DROID~\cite{khazatsky2024droid} & 4.82 & Utokyo Pr2 Tabletop~\cite{oh2023pr2utokyodatasets} & 0.04 \\
    Roboset~\cite{kumar2023robohive} & 3.21 & Utokyo Xarm Pap~\cite{matsushima2023weblab} & 0.04 \\
    FMB Dataset~\cite{luo2023fmb} & 1.50 & CMU Stretch~\cite{mendonca2023structured} & 0.02 \\
    Taco Play~\cite{rosetebeas2022latent,mees2023grounding} & 1.26 & DLR Sara Grid Clamp~\cite{padalkar2023guided} & 0.02 \\
    RoboTurk~\cite{DBLP:journals/corr/abs-1811-02790} & 0.70 & Utokyo Pr2 Fridge~\cite{oh2023pr2utokyodatasets} & 0.01 \\
    Berkeley Autolab Ur5~\cite{BerkeleyUR5Website} & 0.35 & --- & ---\\ 
    \bottomrule
\end{tabular}
\end{table}

\subsection{Downstream Task Descriptions and Success Criteria}
\label{ap:task_protocols}

\textbf{Task 1: Unscrew Bottle Cap.}
The robot first stabilizes a bottle using its left gripper while the right gripper performs repeated rotational motions to unscrew the bottle cap. After the cap is loosened, the right gripper places it onto the tabletop. A trial is considered successful if the robot completes the cap-unscrewing motion with the right gripper.

\textbf{Task 2: Organize Box.}
The robot organizes multiple tabletop objects into designated compartments of a toolbox using both grippers. After placing all objects, the robot closes the intermediate toolbox cover by grasping and repositioning it to the correct location. A trial is considered successful if all objects are placed into their designated compartments and the intermediate toolbox cover is correctly grasped.

\textbf{Task 3: Sort Fruits.}
The left gripper stabilizes a two-layer storage basket while the right gripper sorts fruits and vegetables into different layers of the basket. Fruits are placed in one layer and vegetables in the other. A trial is considered successful only if all objects are placed inside the basket and each object is sorted into its designated layer.

\textbf{Task 4: Put Items to Bag and Zip.}
The robot first picks up a tabletop object and places it into a backpack. It then grasps the right side of the backpack with one gripper while the other gripper grasps the zipper and closes the bag. A trial is considered successful if the object is successfully placed inside the backpack and the zipper is fully closed.

\textbf{Task 5: Pour Water.}
A dexterous hand grasps a bottle and pours its contents into a target cup specified by the instruction. To ensure safe and repeatable evaluation, the bottle cap remains closed during testing, and the task focuses on the pouring motion rather than liquid transfer. A trial is considered successful if the dexterous hand successfully grasps the bottle and executes the complete pouring motion.

\textbf{Task 6: Grasp with a Clamp.}
The dexterous hand manipulates a food clamp to grasp a meat patty, lifts it from the workspace, and transfers it onto a bread bun. A trial is considered successful if the meat patty is successfully transferred and placed on top of the bread bun.

\begin{table*}[t]
\centering
\small
\setlength{\tabcolsep}{4pt}
\renewcommand{\arraystretch}{1.15}
\caption{Statistics of OOL Glove human-hand demonstrations by task type and duration.}
\label{tab:ool_glove_data}
\begin{tabular}{lrrrp{0.37\textwidth}}
\toprule
\textbf{Category} & \textbf{Frames} & \textbf{Duration} & \textbf{Ratio} & \textbf{Included Subcategories} \\
\midrule
Household tasks & 98,680,680 & 913.7 h & 45.7\% & organize kitchen, tidy desk shoe cabinet, refrigerator, storage shelf, set up dining table, prepare medication, clean desks, cabinets, toilet, tableware and items; chop vegetables; trim leaves; make tea or coffee; operate microwave, refrigerator, and coffee machine... \\
Precise tasks & 3,504,600 & 32.5 h & 1.6\% & plug/unplug chargers, plug in mosquito repeller, plug in electric kettle... \\
Deformable tasks & 106,519,320 & 986.3 h & 49.3\% & organize bedroom, tidy bed, tidy sofa, tidy pillowcases, organize clothes and wardrobe, change bedding, wash clothes... \\
Mobile manipulation tasks & 7,295,400 & 67.6 h & 3.4\% & set up meeting room, set up tea room... \\
\bottomrule
\end{tabular}
\label{tab:ooldata}
\end{table*}

\section{Additional Quantitative Results} 
\label{ap:AQuan}

\subsection{Effectiveness of OOL Glove Data}
\label{ap:data_paradigm}

We further evaluate the quality of OOL Glove data against other data collection paradigms under the same LaST-HD training recipe. As shown in the right part of Fig.~\ref{fig:ablation}(b), across three generalization scenarios, we compare 60 OOL Glove demonstrations with 60 vision-based Bare-hand demonstrations~\citep{zhang2025hawor}, as well as real-robot data using either 12 trajectories per scenario (Real-12), which matches the collection time of 60 human-hand demonstrations, or 60 trajectories per scenario (Real-60). Implementation details are provided in Appendix~\ref{ap:BID}. OOL Glove data achieves 73\%, consistently outperforming bare-hand demonstrations (63\%), indicating higher-quality interaction data. We further reproduce the UMI data-collection paradigm following its official setup~\cite{chi2024universal}, capturing three synchronized RGB views per demonstration: one head-mounted view and two wrist-mounted views. Trained under the same recipe, LaST-HD reaches 65\% with real-world and UMI data, outperforming the bare-hand baseline but still falling short of our OOL Glove setup. Although UMI provides relatively reliable demonstrations, its cross-embodiment reusability remains limited: as a two-finger parallel gripper, it captures only two-finger interactions and therefore cannot be retargeted to high-DoF dexterous hands. Meanwhile, under the same collection time, our method outperforms real-robot data (Real-12) by 13\%, while remaining comparable to the Real-60 setting. This shows that our latent alignment strategy enables efficient transfer of human interaction data into robot action learning. We also find that placing the wrist camera near the thumb-index web space outperforms the palm-view setup (67\%). The exact success rates of all paradigms are summarized in Table~\ref{tab:training_data}, and additional ablations on LaST-HD hyperparameters are presented in the following subsections.

\begin{table}[htbp]
    \centering
    \small
    \setlength{\tabcolsep}{5pt}
    \caption{Effect of training data source on LaST-HD manipulation performance.}
    \begin{tabular}{l c c c c c}
        \toprule
        \textbf{Training data} & \textbf{Real-60} & \textbf{Real-12} & \textbf{Bare hand} & \textbf{UMI} & \textbf{Palm view}   \\
        \midrule
        Success Rate & 0.75 & 0.60 & 0.63 & 0.65 & 0.67 \\
        \bottomrule
    \end{tabular}
    \label{tab:training_data}
\end{table}

\subsection{Additional Ablation Studies}

Unless otherwise specified, all ablations are conducted on the dual-arm \textit{Sort Fruits} task. We choose this task because it requires non-trivial bimanual coordination and multi-step object manipulation, making it highly representative of LaST-HD's reasoning and control capabilities. At the same time, compared with dexterous-hand tasks, it avoids introducing additional confounding factors arising from high-dimensional hand control, providing a cleaner evaluation of the proposed design choices.

\begin{table}[htbp]
    \centering
    \caption{Effect of denoising steps for latent target extraction on the manipulation performance.}
    \begin{tabular}{l c c c}
        \toprule
        \textbf{Denoising steps} & \textbf{2} & \textbf{5} & \textbf{10}  \\
        \midrule
        Success Rate & 0.73 & 0.72 & 0.76  \\
        \bottomrule
    \end{tabular}
    \label{tab:noise}
\end{table}

\textbf{Effect of world-model denoising steps.}
We first investigate how the denoising step used for latent target extraction affects the final policy performance. Specifically, when constructing latent ground-truth targets with the action-conditioned world model, we extract latent features after 2, 5, and 10 denoising steps, respectively. As shown in Table~\ref{tab:noise}, all three settings achieve similar downstream success rates, indicating that the latent supervision remains largely stable across different denoising depths. Given the negligible performance difference, we adopt 2 denoising steps throughout all experiments. Although the world model is only used offline for latent target generation and is not involved during LaST-HD VLA model inference, reducing the denoising steps significantly improves the efficiency of training dataset construction.

\begin{table}[htbp]
    \centering
    \caption{Impact of shared latent length on LaST-HD manipulation performance.}
    \begin{tabular}{l c c c c c}
        \toprule
        \textbf{Latent length} & \textbf{2} & \textbf{4} & \textbf{8} & \textbf{12} & \textbf{16}   \\
        \midrule
        Success Rate & 0.67 & 0.73 & 0.67 & 0.70  &  0.78 \\
        \bottomrule
    \end{tabular}
    \label{tab:latent length}
\end{table}

\textbf{Effect of LaST-HD's shared latent length.}
We further study the impact of the shared latent length in LaST-HD. Specifically, we evaluate latent lengths of 2, 4, 8, 12, and 16 tokens while keeping all other training configurations fixed. 
As shown in Table~\ref{tab:latent length}, latent lengths below 12 tokens achieve comparable performance. When the latent length is increased to 16, we observe an obvious improvement in task success rate.
However, because latent generation inherits the native autoregressive decoding paradigm of LLMs, increasing the latent sequence length directly increases inference latency. Considering the trade-off between performance and efficiency, we adopt a latent length of 4 in all experiments.

\begin{figure*}[t]
    \centering
    \includegraphics[width=\textwidth]{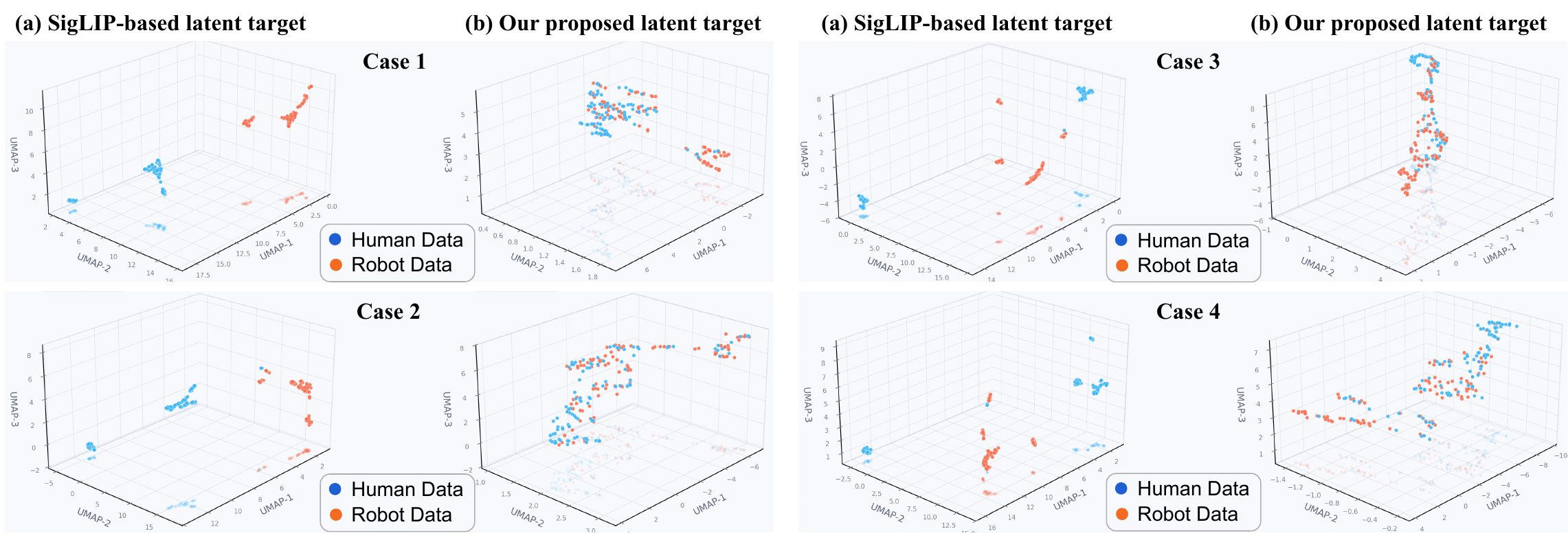} 
    \vspace{-0.2cm}
    \caption{
    UMAP visualization of LaST-HD latent tokens generated by different model variants. We compare latent representations learned from (a) SigLIP-based latent targets and (b) our action-conditioned world-model latent targets.
    }
    \label{fig:apumap}
    \vspace{-0.2cm}
\end{figure*}

\section{Additional Qualitative Results}
\label{ap:AQual}
\subsection{Additional UMAP Visualization}

To better understand how LaST-HD aligns human-hand demonstrations with robot trajectories, we provide additional UMAP visualizations of the learned latent representations in Figure~X. 
Following prior work~\citep{lei2026mechanistic}, we project the latent tokens produced by the reasoning expert into a three-dimensional space using UMAP~\citep{mcinnes2018umap}.
Specifically, we separately collect latent representations from human-hand inputs and robot inputs under the same task and visualize them jointly.

As shown in Figure~\ref{fig:apumap}, LaST-HD consistently produces structured alignment between human-hand and robot trajectories belonging to the same task. Rather than forming two isolated clusters corresponding to human hand and robot, the latent representations exhibit overlapping geometric structures. This observation suggests that the proposed shared latent reasoning space successfully bridges the embodiment gap between human demonstrations and robot executions. Compared with SigLIP-based latent targets, the action-conditioned world-model supervision encourages latent representations to capture shared physical dynamics and task-relevant interaction patterns. As a result, human-hand demonstrations and robot trajectories become more closely aligned in the latent space, enabling more effective transfer of manipulation knowledge during mixed human-robot training. This structured alignment directly contributes to the strong data efficiency of LaST-HD when leveraging human-hand demonstrations.

\begin{figure*}[t]
    \centering
    \includegraphics[width=0.7\textwidth]{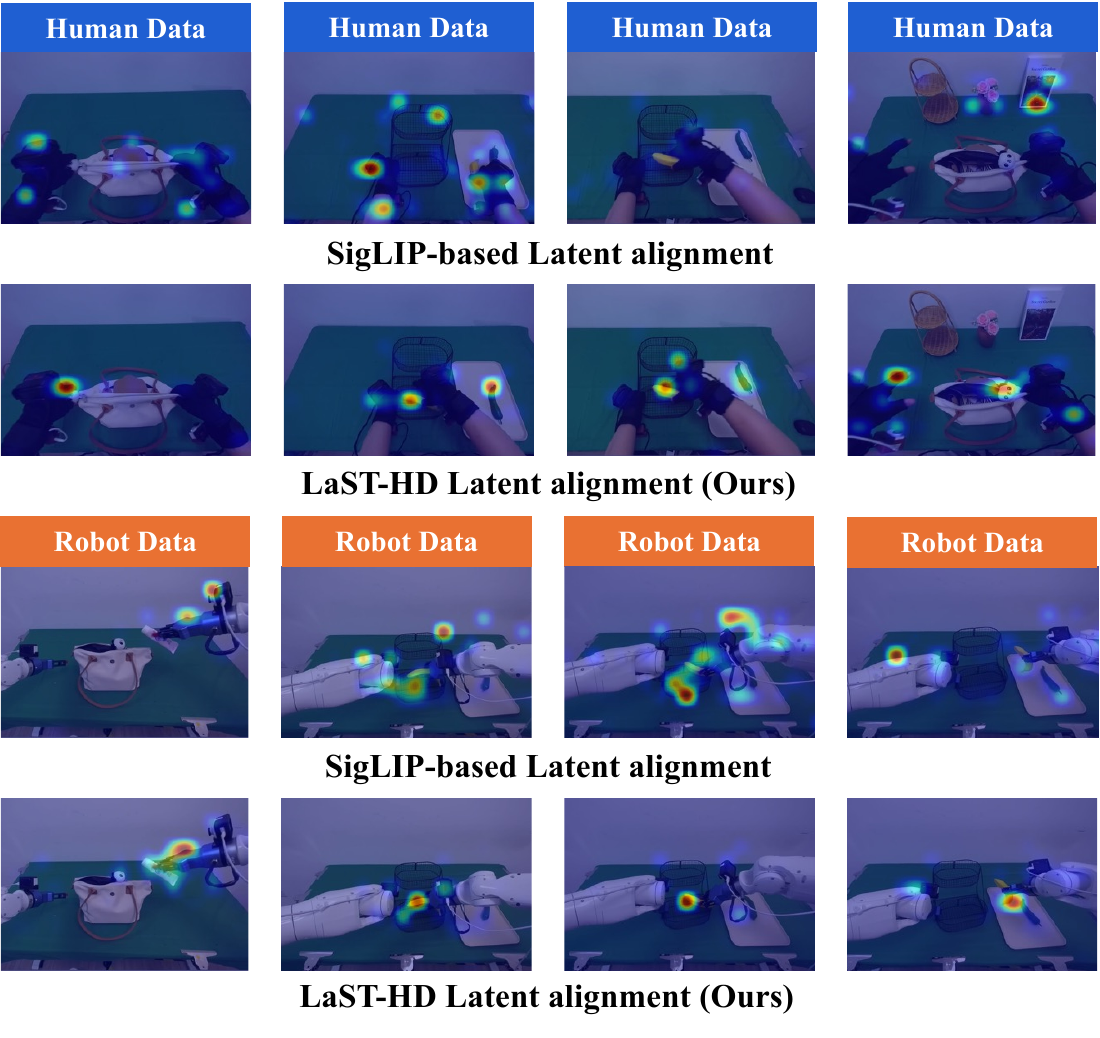} 
    \vspace{-0.1cm}
    \caption{
    Visualization of LaST-HD output-token-to-visual-token attention maps. 
    }
    \label{fig:apatt}
    \vspace{-0.2cm}
\end{figure*}

\subsection{Additional Attention Map Visualization.}
To explicitly demonstrate how the proposed latent alignment facilitates action generation, we visualize the attention maps between LaST-HD output tokens and visual tokens in Figure~\ref{fig:apatt}.
The resulting attention distributions are projected back to the image space for visualization.

The attention maps further support the UMAP analysis by showing that LaST-HD focuses primarily on embodiment-object interactions and manipulation-relevant physical dynamics. Across both human-hand and robot observations, the latent tokens consistently attend to manipulated objects, contact regions, and interaction points, while suppressing irrelevant background distractions. In contrast, SigLIP-based latent supervision tends to allocate attention more broadly across the scene and is less concentrated on the manipulated object.
These visual findings elucidate the mechanism behind our shared latent reasoning: by constraining latent representations to capture task-relevant dynamics rather than domain-specific visual appearances, LaST-HD establishes a domain-invariant reasoning space that effectively harnesses human demonstrations to enhance robot policy learning.

\subsection{Additional Visualizations of Robot Executions and OOL Glove Data}
\begin{figure*}[t]
    \centering
    \includegraphics[width=1.0\textwidth]{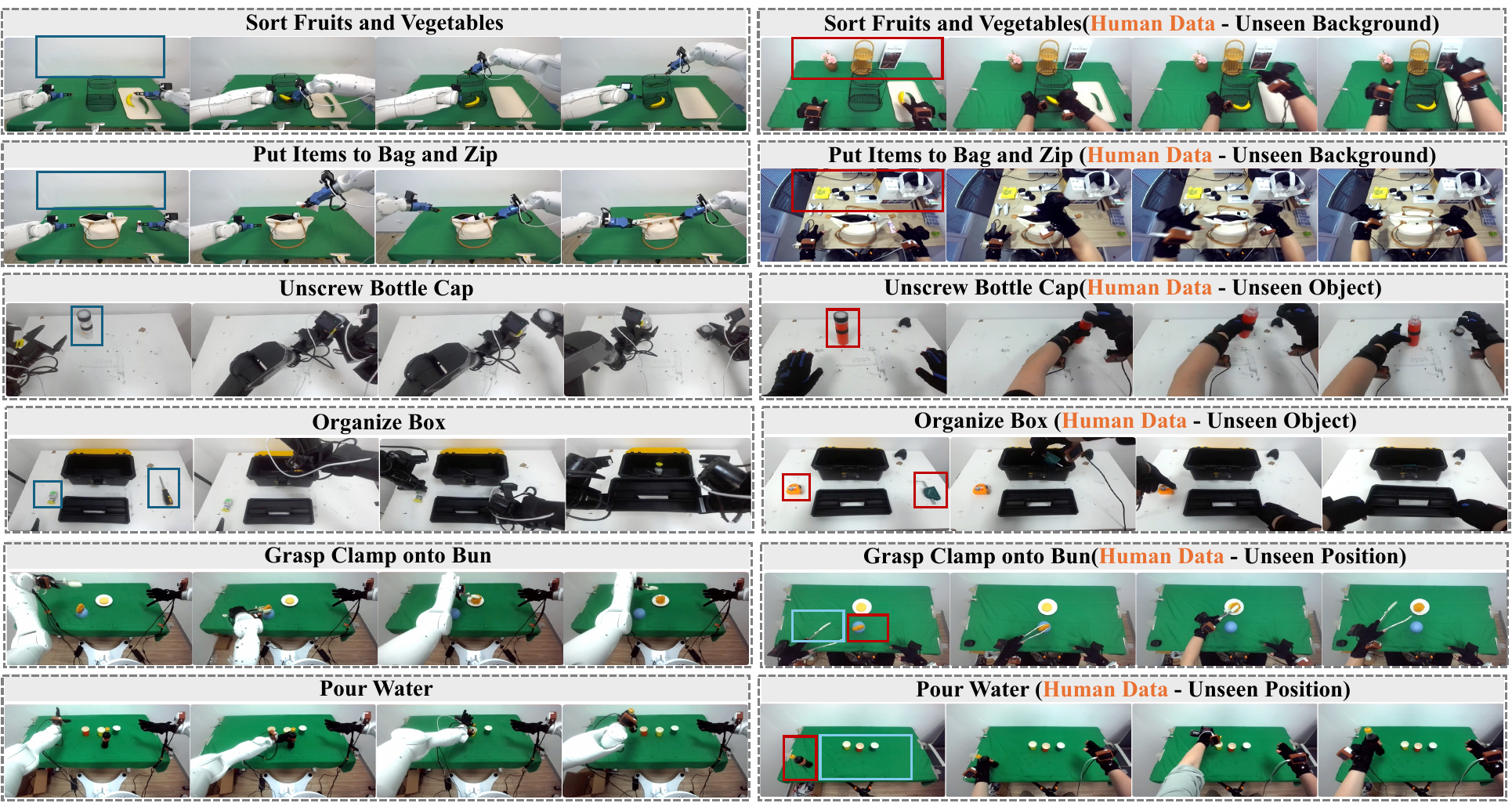} 
    \vspace{-0.3cm}
    \caption{
    Additional visualizations of robot executions and OOL Glove Data.
    }
    \label{fig:add_viz}
    \vspace{-0.2cm}
\end{figure*}

We provide additional visualizations of robot execution progress and OOL Glove data in unseen scenarios. The left side of Figure~\ref{fig:add_viz} shows sequential robot executions across several manipulation tasks, while the right side presents corresponding human demonstrations collected with the OOL Glove under unseen conditions, including unseen backgrounds, manipulated object, and object positions. 
Blue boxes denote in-domain configurations, whereas red boxes highlight unseen configurations. These visualizations demonstrate that low-cost human-hand data, combined with our proposed training recipe, enables robots to reliably execute diverse real-world tasks. Video demonstrations are provided in the supplementary material.

\section{Baseline Implementation Details}
\label{ap:BID}

\subsection{Model Baselines}

\textbf{$\pi_{0.5}$~\cite{intelligence2025pi05visionlanguageactionmodelopenworld}}
is a PaliGemma-based~\cite{beyer2024paligemma} vision-language-action (VLA) model that employs flow matching for action generation. We adopt full-parameter fine-tuning of the entire model during downstream post-training. Following its official open-source implementation, all input images are resized to $224 \times 224$. We keep this resolution unchanged, since modifying the image input size may disrupt the visual representations learned during pre-training.

\textbf{Cosmos-Policy~\cite{kim2026cosmos}}
adapts a pretrained Cosmos-Predict video foundation model into a robot policy through post-training on robot demonstrations. It represents actions, future states, and value estimates as latent frames within the video diffusion process, enabling visuomotor control. Following the official input image resolution, we resize all input images to $224 \times 224$.

\textbf{LaST$_0$~\cite{liu2026last}}
is a Janus-Pro-based~\cite{chen2025janus} VLA model that introduces a Mixture-of-Transformers architecture for latent spatio-temporal Chain-of-Thought reasoning over future visual, 3D, and proprioceptive states. Following the official setup, all input images are resized to $384 \times 384$.

For all model baselines, when training on high-DoF dexterous-hand tasks, we directly modify the action output dimension to match the corresponding dexterous action space while keeping the remaining training configurations unchanged.

\subsection{Data Collection Baselines}

\textbf{UMI~\cite{chi2024universal}}
collects data using portable hand-held grippers equipped with cameras, allowing humans to directly demonstrate manipulation skills in the wild. Following official setups, we use three GoPro cameras to capture multi-view visual observations during data collection. The demonstrations include visual observations and 6-DoF gripper trajectories, which are then converted into robot-executable action data for policy learning.

\textbf{Hawor~\cite{zhang2025hawor}}
collects human-hand demonstrations from egocentric videos and reconstructs world-space hand trajectories through hand-pose estimation and camera-motion tracking. By combining adaptive egocentric SLAM with motion reconstruction, HawOR produces temporally consistent hand trajectories that can be retargeted to robot embodiments for imitation learning.

\section{Failure Case Analysis}
\label{ap:FCA}

\begin{figure*}[t]
    \centering
    \includegraphics[width=1.0\textwidth]{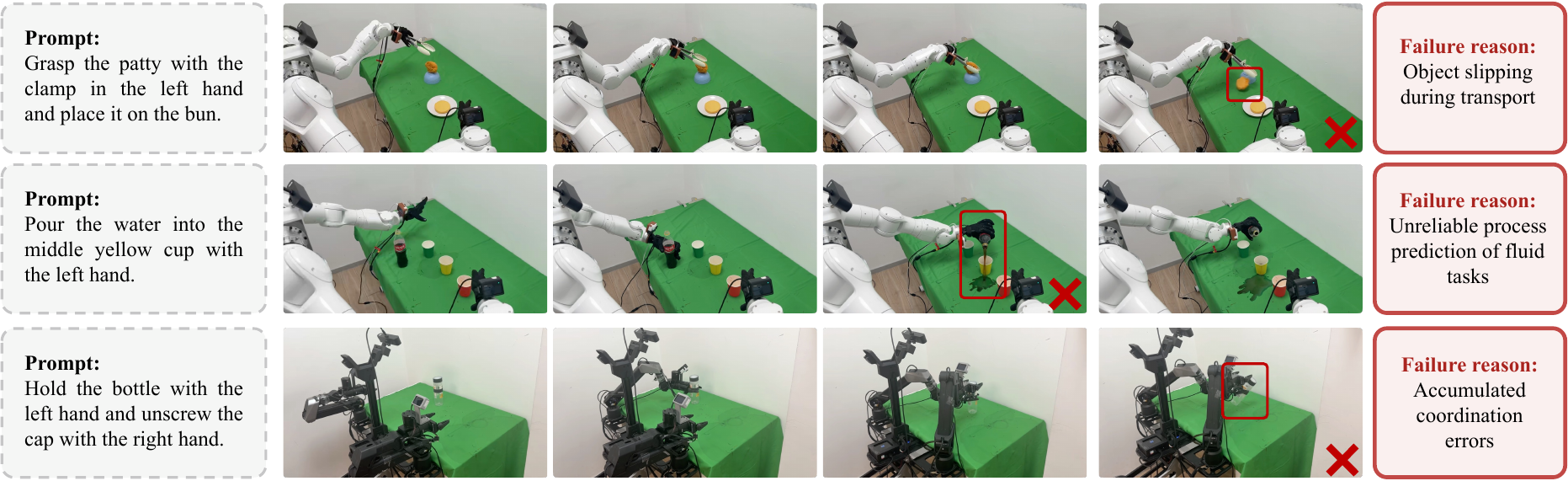} 
    \caption{
    Representative failure cases observed during real-world deployment of LaST-HD.
    }
    \label{fig:failure}
\end{figure*}

As shown in Figure~\ref{fig:failure}, although LaST-HD achieves strong performance across different embodiments and tasks, execution failures still occur in a small number of complex scenarios, as shown in Figure X. These failures typically arise from hardware execution errors or spatial coordination challenges. The main failure cases are summarized below:

\textbf{1) Object slipping during transport:} For the dexterous-hand \textit{Grasp Clamp onto Bun} task, the model can correctly plan the approach and lifting trajectory, but the selected grasp point on the tool may occasionally be suboptimal. In addition, the dexterous hand may apply imprecise contact forces, leading to unstable grasping or placement. As a result, the object is not firmly held and may slip out during transport, causing task failure. Although our model has a certain degree of closed-loop recovery capability and may attempt to re-clamp the object, this substantially reduces task efficiency. Moreover, if the object falls into an unrecoverable state, the task ultimately fails.

\textbf{2) Unreliable process prediction of fluid tasks:} While the model is able to approach the bottle and initiate a stable pouring motion, the dynamics of flowing liquids are highly stochastic and difficult to predict precisely. As a result, the model may sometimes fail to determine whether the water is still being poured, and spilling may occur, leading to task failure. Accurately predicting fluid behavior remains a long-standing challenge for general manipulation methods. In future work, we plan to incorporate fluid-aware perception and prediction modules to improve the precision of liquid-pouring tasks.

\textbf{3) Accumulated coordination errors:} Even though the model correctly coordinates the two arms to stabilize the bottle and unscrew the cap, maintaining continuous physical contact between the arms is a highly delicate task. Small errors in their relative poses can gradually accumulate during execution. As a result, an inaccurate left-hand grasp may tilt the bottle, making it difficult for the right hand to complete the cap-unscrewing motion. In future work, we plan to incorporate tighter dual-arm coordination strategies to mitigate such accumulated errors.

%% file: example.bib
@article{luo2023diffusion,
  title={Diffusion hyperfeatures: Searching through time and space for semantic correspondence},
  author={Luo, Grace and Dunlap, Lisa and Park, Dong Huk and Holynski, Aleksander and Darrell, Trevor},
  journal={Advances in Neural Information Processing Systems},
  volume={36},
  pages={47500--47510},
  year={2023}
}

@article{mcinnes2018umap,
  title={Umap: Uniform manifold approximation and projection for dimension reduction},
  author={McInnes, Leland and Healy, John and Melville, James},
  journal={arXiv preprint arXiv:1802.03426},
  year={2018}
}

@article{kim2024openvla,
  title={OpenVLA: An Open-Source Vision-Language-Action Model},
  author={Kim, Moo Jin and Pertsch, Karl and Karamcheti, Siddharth and Xiao, Ted and Balakrishna, Ashwin and Nair, Suraj and Rafailov, Rafael and Foster, Ethan and Lam, Grace and Sanketi, Pannag and others},
  journal={arXiv preprint arXiv:2406.09246},
  year={2024}
}

@article{black2024pi_0,
  title={pi0: A Vision-Language-Action Flow Model for General Robot Control},
  author={Black, Kevin and Brown, Noah and Driess, Danny and Esmail, Adnan and Equi, Michael and Finn, Chelsea and Fusai, Niccolo and Groom, Lachy and Hausman, Karol and Ichter, Brian and others},
  journal={arXiv preprint arXiv:2410.24164},
  year={2024}
}

@article{liu2025hybridvla,
  title={HybridVLA: Collaborative Diffusion and Autoregression in a Unified Vision-Language-Action Model},
  author={Liu, Jiaming and Chen, Hao and An, Pengju and Liu, Zhuoyang and Zhang, Renrui and Gu, Chenyang and Li, Xiaoqi and Guo, Ziyu and Chen, Sixiang and Liu, Mengzhen and others},
  journal={arXiv preprint arXiv:2503.10631},
  year={2025}
}

@String(CVPR= {IEEE Conf. Comput. Vis. Pattern Recog.})

@String(ICCV= {Int. Conf. Comput. Vis.})

@String(ECCV= {Eur. Conf. Comput. Vis.})

@String(AAAI = {AAAI})

@String(CVPR  = {CVPR})

@String(ICCV  = {ICCV})

@String(ECCV  = {ECCV})

@misc{open_x_embodiment_rt_x_2023,
title={Open {X-E}mbodiment: Robotic Learning Datasets and {RT-X} Models},
author = {{Open X-Embodiment Collaboration} and Abhishek Padalkar and Acorn Pooley and others},
howpublished  = {\url{https://arxiv.org/abs/2310.08864}},
year = {2023},
}

@article{khazatsky2024droid,
    title   = {DROID: A Large-Scale In-The-Wild Robot Manipulation Dataset},
    author  = {Alexander Khazatsky and Karl Pertsch and Suraj Nair and Ashwin Balakrishna and Sudeep Dasari and Siddharth Karamcheti and Soroush Nasiriany and Mohan Kumar Srirama and Lawrence Yunliang Chen and Kirsty Ellis and others},
    year    = {2024},
}

@inproceedings{zitkovich2023rt,
  title={Rt-2: Vision-language-action models transfer web knowledge to robotic control},
  author={Zitkovich, Brianna and Yu, Tianhe and Xu, Sichun and Xu, Peng and Xiao, Ted and Xia, Fei and Wu, Jialin and Wohlhart, Paul and Welker, Stefan and Wahid, Ayzaan and others},
  booktitle={7th Annual Conference on Robot Learning},
  year={2023}
}

@article{kalashnikov2018qt,
  title={{QT}-{O}pt: Scalable deep reinforcement learning for vision-based robotic manipulation},
  author={Kalashnikov, Dmitry and Irpan, Alex and Pastor, Peter and Ibarz, Julian and Herzog, Alexander and Jang, Eric and Quillen, Deirdre and Holly, Ethan and Kalakrishnan, Mrinal and Vanhoucke, Vincent and others},
  journal={arXiv preprint arXiv:1806.10293},
  year={2018}
}

@misc{walke2023bridgedata,
      title={BridgeData V2: A Dataset for Robot Learning at Scale}, 
      author={Homer Walke and Kevin Black and Abraham Lee and Moo Jin Kim and Max Du and Chongyi Zheng and Tony Zhao and Philippe Hansen-Estruch and Quan Vuong and Andre He and Vivek Myers and Kuan Fang and Chelsea Finn and Sergey Levine},
      year={2023},
      eprint={2308.12952},
      archivePrefix={arXiv},
      primaryClass={cs.RO}
}

@inproceedings{rosetebeas2022latent,
author = {Erick Rosete-Beas and Oier Mees and Gabriel Kalweit and Joschka Boedecker and Wolfram Burgard},
title = {Latent Plans for Task Agnostic Offline Reinforcement Learning},
booktitle = {Proceedings of the 6th Conference on Robot Learning (CoRL)},
year = {2022}
}

@inproceedings{mees2023grounding,
  title={Grounding  Language  with  Visual  Affordances  over  Unstructured  Data},
  author={Oier Mees and Jessica Borja-Diaz and Wolfram Burgard},
  booktitle = {Proceedings of the IEEE International Conference on Robotics and Automation (ICRA)},
  year={2023},
  address = {London, UK}
}

@misc{dass2023jacoplay,
  author = {Dass, Shivin and Yapeter, Jullian and Zhang, Jesse and Zhang, Jiahui
            and Pertsch, Karl and Nikolaidis, Stefanos and Lim, Joseph J.},
  title = {{CLVR} Jaco Play Dataset},
  url = {https://github.com/clvrai/clvr_jaco_play_dataset},
  version = {1.0.0},
  year = {2023}
}

@article{DBLP:journals/corr/abs-1811-02790,
  author       = {Ajay Mandlekar and
                  Yuke Zhu and
                  Animesh Garg and
                  Jonathan Booher and
                  Max Spero and
                  Albert Tung and
                  Julian Gao and
                  John Emmons and
                  Anchit Gupta and
                  Emre Orbay and
                  Silvio Savarese and
                  Li Fei{-}Fei},
  title        = {{R}obo{T}urk: {A} Crowdsourcing Platform for Robotic Skill Learning through
                  Imitation},
  journal      = {CoRR},
  volume       = {abs/1811.02790},
  year         = {2018},
  eprinttype    = {arXiv},
  eprint       = {1811.02790},
  bibsource    = {dblp computer science bibliography, https://dblp.org}
}

@misc{BerkeleyUR5Website,
  title = {Berkeley {UR5} Demonstration Dataset},
  author = {Lawrence Yunliang Chen and Simeon Adebola and Ken Goldberg},
  howpublished = {\url{https://sites.google.com/view/berkeley-ur5/home}},
}

@misc{zhou2023train,
      title={Train Offline, Test Online: A Real Robot Learning Benchmark}, 
      author={Gaoyue Zhou and Victoria Dean and Mohan Kumar Srirama and Aravind Rajeswaran and Jyothish Pari and Kyle Hatch and Aryan Jain and Tianhe Yu and Pieter Abbeel and Lerrel Pinto and Chelsea Finn and Abhinav Gupta},
      year={2023},
      eprint={2306.00942},
      archivePrefix={arXiv},
      primaryClass={cs.RO}
}

@article{lynch2023interactive,
  title={Interactive language: Talking to robots in real time},
  author={Lynch, Corey and Wahid, Ayzaan and Tompson, Jonathan and Ding, Tianli and Betker, James and Baruch, Robert and Armstrong, Travis and Florence, Pete},
  journal={IEEE Robotics and Automation Letters},
  year={2023},
  publisher={IEEE}
}

@inproceedings{
kumar2023robohive,
title={RoboHive: A Unified Framework for Robot Learning},
author={Vikash Kumar and Rutav Shah and Gaoyue Zhou and Vincent Moens and Vittorio Caggiano and Abhishek Gupta and Aravind Rajeswaran},
booktitle={Thirty-seventh Conference on Neural Information Processing Systems Datasets and Benchmarks Track},
year={2023}
}

@inproceedings{vc2023,
        title         = {Where are we in the search for an Artificial Visual Cortex for Embodied Intelligence?}, 
        author        = {Arjun Majumdar and Karmesh Yadav and Sergio Arnaud and Yecheng Jason Ma and Claire Chen and
                         Sneha Silwal and Aryan Jain and Vincent-Pierre Berges and Pieter Abbeel and Jitendra Malik and
                         Dhruv Batra and Yixin Lin and Oleksandr Maksymets and Aravind Rajeswaran and Franziska Meier},
        year          = {2023},
        eprint        = {2303.18240},
        archivePrefix = {arXiv},
        primaryClass  = {cs.CV}
    }

@inproceedings{ma2022vip,
  title={VIP: Towards Universal Visual Reward and Representation via Value-Implicit Pre-Training},
  author={Ma, Yecheng Jason and Sodhani, Shagun and Jayaraman, Dinesh and Bastani, Osbert and Kumar, Vikash and Zhang, Amy},
  booktitle={The Eleventh International Conference on Learning Representations},
  year={2022}
}

@inproceedings{jang2022bc,
  title={Bc-z: Zero-shot task generalization with robotic imitation learning},
  author={Jang, Eric and Irpan, Alex and Khansari, Mohi and Kappler, Daniel and Ebert, Frederik and Lynch, Corey and Levine, Sergey and Finn, Chelsea},
  booktitle={Conference on Robot Learning},
  pages={991--1002},
  year={2022},
  organization={PMLR}
}

@article{belkhale2023hydra,
 title={HYDRA: Hybrid Robot Actions for Imitation Learning},
 author={Belkhale, Suneel and Cui, Yuchen and Sadigh, Dorsa},
 journal={arxiv},
 year={2023}
}

@misc{luo2023fmb,
  title = {{FMB}: A Functional Manipulation Benchmark for Generalizable Robotic Learning},
  author={Jianlan Luo and Charles Xu and Fangchen Liu and Liam Tan and Zipeng Lin and Jeffrey Wu and Pieter Abbeel and Sergey Levine},
  year={2023},
  howpublished = {\url{https://functional-manipulation-benchmark.github.io}},
}

@article{cui2022play,
  title   = {From Play to Policy: Conditional Behavior Generation from Uncurated Robot Data},
  author  = {Cui, Zichen Jeff and Wang, Yibin and Shafiullah, Nur Muhammad Mahi and Pinto, Lerrel},
  journal = {arXiv preprint arXiv:2210.10047},
  year    = {2022}
}

@inproceedings{heo2023furniturebench,
  title={FurnitureBench: Reproducible Real-World Benchmark for Long-Horizon Complex Manipulation},
  author={Minho Heo and Youngwoon Lee and Doohyun Lee and Joseph J. Lim},
  booktitle={Robotics: Science and Systems},
  year={2023}
}

@inproceedings{bahl2023affordances,
  title={Affordances from Human Videos as a Versatile Representation for Robotics},
  author={Bahl, Shikhar and Mendonca, Russell and Chen, Lili and Jain, Unnat and Pathak, Deepak},
  booktitle={CVPR},
  year={2023}
}

@article{mendonca2023structured,
  title={Structured World Models from Human Videos},
  author={Mendonca, Russell and Bahl, Shikhar and Pathak, Deepak},
  journal={CoRL},
  year={2023}
}

@inproceedings{dasari2020robonet,
  title={RoboNet: Large-Scale Multi-Robot Learning},
  author={Dasari, Sudeep and Ebert, Frederik and Tian, Stephen and Nair, Suraj and Bucher, Bernadette and Schmeckpeper, Karl and Singh, Siddharth and Levine, Sergey and Finn, Chelsea},
  booktitle={Conference on Robot Learning},
  pages={885--897},
  year={2020},
  organization={PMLR}
}

@misc{shafiullah2023dobbe,
    title={On Bringing Robots Home}, 
    author={Nur Muhammad Mahi Shafiullah and Anant Rai and Haritheja Etukuru and Yiqian Liu and Ishan Misra and Soumith Chintala and Lerrel Pinto},
    year={2023},
    eprint={2311.16098},
    archivePrefix={arXiv},
    primaryClass={cs.RO}
}

@article{bahl2022human,
  title={Human-to-robot imitation in the wild},
  author={Bahl, Shikhar and Gupta, Abhinav and Pathak, Deepak},
  journal={arXiv preprint arXiv:2207.09450},
  year={2022}
}

@article{chen2021learning,
  title={Learning Generalizable Robotic Reward Functions from" In-The-Wild" Human Videos},
  author={Chen, Annie S and Nair, Suraj and Finn, Chelsea},
  journal={RSS},
  year={2021}
}

@inproceedings{damen2018scaling,
  title={Scaling egocentric vision: The epic-kitchens dataset},
  author={Damen, Dima and Doughty, Hazel and Maria Farinella, Giovanni and Fidler, Sanja and Furnari, Antonino and Kazakos, Evangelos and Moltisanti, Davide and Munro, Jonathan and Perrett, Toby and Price, Will and others},
  booktitle={Proceedings of the European Conference on Computer Vision (ECCV)},
  pages={720--736},
  year={2018}
}

@inproceedings{grauman2022ego4d,
  title={Ego4d: Around the world in 3,000 hours of egocentric video},
  author={Grauman, Kristen and Westbury, Andrew and Byrne, Eugene and Chavis, Zachary and Furnari, Antonino and Girdhar, Rohit and Hamburger, Jackson and Jiang, Hao and Liu, Miao and Liu, Xingyu and others},
  booktitle={Proceedings of the IEEE/CVF Conference on Computer Vision and Pattern Recognition},
  pages={18995--19012},
  year={2022}
}

@inproceedings{ebert2022bridge,
  title={Bridge data: Boosting generalization of robotic skills with cross-domain datasets},
  author={Ebert, Frederik and Yang, Yanlai and Schmeckpeper, Karl and Bucher, Bernadette and Georgakis, Georgios and Daniilidis, Kostas and Finn, Chelsea and Levine, Sergey},
  booktitle={RSS},
  year={2022}
}

@inproceedings{nair2022r3m,
  title={R3m: A universal visual representation for robot manipulation},
  author={Nair, Suraj and Rajeswaran, Aravind and Kumar, Vikash and Finn, Chelsea and Gupta, Abhinav},
  booktitle={CoRL},
  year={2022}
}

@inproceedings{rt12022arxiv,
    title={RT-1: Robotics Transformer for Real-World Control at Scale},
    author={Anthony	Brohan and  Noah Brown and  Justice Carbajal and  Yevgen Chebotar and  Joseph Dabis and  Chelsea Finn and  Keerthana Gopalakrishnan and  Karol Hausman and  Alex Herzog and  Jasmine Hsu and  Julian Ibarz and others},
    booktitle={arXiv preprint arXiv:2212.06817},
    year={2022}
}

@inproceedings{zhai2023siglip,
  author = {Xiaohua Zhai and Basil Mustafa and Alexander Kolesnikov and Lucas Beyer},
  booktitle = {International Conference on Computer Vision (ICCV)},
  title = {Sigmoid Loss for Language Image Pre-Training},
  year = {2023},
}

@inproceedings{o2024open,
  title={Open x-embodiment: Robotic learning datasets and rt-x models: Open x-embodiment collaboration 0},
  author={O’Neill, Abby and Rehman, Abdul and Maddukuri, Abhiram and Gupta, Abhishek and Padalkar, Abhishek and Lee, Abraham and Pooley, Acorn and Gupta, Agrim and Mandlekar, Ajay and Jain, Ajinkya and others},
  booktitle={2024 IEEE International Conference on Robotics and Automation (ICRA)},
  pages={6892--6903},
  year={2024},
  organization={IEEE}
}

@article{chi2024universal,
  title={Universal manipulation interface: In-the-wild robot teaching without in-the-wild robots},
  author={Chi, Cheng and Xu, Zhenjia and Pan, Chuer and Cousineau, Eric and Burchfiel, Benjamin and Feng, Siyuan and Tedrake, Russ and Song, Shuran},
  journal={arXiv preprint arXiv:2402.10329},
  year={2024}
}

@inproceedings{zhao2025cot,
  title={Cot-vla: Visual chain-of-thought reasoning for vision-language-action models},
  author={Zhao, Qingqing and Lu, Yao and Kim, Moo Jin and Fu, Zipeng and Zhang, Zhuoyang and Wu, Yecheng and Li, Zhaoshuo and Ma, Qianli and Han, Song and Finn, Chelsea and others},
  booktitle={Proceedings of the Computer Vision and Pattern Recognition Conference},
  pages={1702--1713},
  year={2025}
}

@article{tang2023emergent,
  title={Emergent correspondence from image diffusion},
  author={Tang, Luming and Jia, Menglin and Wang, Qianqian and Phoo, Cheng Perng and Hariharan, Bharath},
  journal={Advances in neural information processing systems},
  volume={36},
  pages={1363--1389},
  year={2023}
}

@article{wen2024diffusion,
  title={Diffusion-VLA: Scaling Robot Foundation Models via Unified Diffusion and Autoregression},
  author={Wen, Junjie and Zhu, Minjie and Zhu, Yichen and Tang, Zhibin and Li, Jinming and Zhou, Zhongyi and Li, Chengmeng and Liu, Xiaoyu and Peng, Yaxin and Shen, Chaomin and others},
  journal={arXiv preprint arXiv:2412.03293},
  year={2024}
}

@article{zhang2025dreamvla,
  title={DreamVLA: a vision-language-action model dreamed with comprehensive world knowledge},
  author={Zhang, Wenyao and Liu, Hongsi and Qi, Zekun and Wang, Yunnan and Yu, Xinqiang and Zhang, Jiazhao and Dong, Runpei and He, Jiawei and Wang, He and Zhang, Zhizheng and others},
  journal={arXiv preprint arXiv:2507.04447},
  year={2025}
}

@article{gao2025adaworld,
  title={Adaworld: Learning adaptable world models with latent actions},
  author={Gao, Shenyuan and Zhou, Siyuan and Du, Yilun and Zhang, Jun and Gan, Chuang},
  journal={arXiv preprint arXiv:2503.18938},
  year={2025}
}

@article{cen2025worldvla,
  title={WorldVLA: Towards Autoregressive Action World Model},
  author={Cen, Jun and Yu, Chaohui and Yuan, Hangjie and Jiang, Yuming and Huang, Siteng and Guo, Jiayan and Li, Xin and Song, Yibing and Luo, Hao and Wang, Fan and others},
  journal={arXiv preprint arXiv:2506.21539},
  year={2025}
}

@article{ye2024latent,
  title={Latent action pretraining from videos},
  author={Ye, Seonghyeon and Jang, Joel and Jeon, Byeongguk and Joo, Sejune and Yang, Jianwei and Peng, Baolin and Mandlekar, Ajay and Tan, Reuben and Chao, Yu-Wei and Lin, Bill Yuchen and others},
  journal={arXiv preprint arXiv:2410.11758},
  year={2024}
}

@inproceedings{wu2024robomind,
  title={Robomind: Benchmark on multi-embodiment intelligence normative data for robot manipulation},
  author={Wu, Kun and Hou, Chengkai and Liu, Jiaming and Che, Zhengping and Ju, Xiaozhu and others},
	booktitle={Robotics: Science and Systems (RSS) 2025}, 
  year={2025},
  publisher={Robotics: Science and Systems Foundation}, 
}

@misc{gu2023maniskill2unifiedbenchmarkgeneralizable,
      title={ManiSkill2: A Unified Benchmark for Generalizable Manipulation Skills}, 
      author={Jiayuan Gu and Fanbo Xiang and Xuanlin Li and Zhan Ling and Xiqiang Liu and Tongzhou Mu and Yihe Tang and Stone Tao and Xinyue Wei and Yunchao Yao and Xiaodi Yuan and Pengwei Xie and Zhiao Huang and Rui Chen and Hao Su},
      year={2023},
      eprint={2302.04659},
      archivePrefix={arXiv},
      primaryClass={cs.RO},
      url={https://arxiv.org/abs/2302.04659}, 
}

@article{chen2025fast,
  title={Fast-in-Slow: A Dual-System Foundation Model Unifying Fast Manipulation within Slow Reasoning},
  author={Chen, Hao and Liu, Jiaming and Gu, Chenyang and Liu, Zhuoyang and Zhang, Renrui and Li, Xiaoqi and He, Xiao and Guo, Yandong and Fu, Chi-Wing and Zhang, Shanghang and others},
  journal={arXiv preprint arXiv:2506.01953},
  year={2025}
}

@article{bjorck2025gr00t,
  title={Gr00t n1: An open foundation model for generalist humanoid robots},
  author={Bjorck, Johan and Casta{\~n}eda, Fernando and Cherniadev, Nikita and Da, Xingye and Ding, Runyu and Fan, Linxi and Fang, Yu and Fox, Dieter and Hu, Fengyuan and Huang, Spencer and others},
  journal={arXiv preprint arXiv:2503.14734},
  year={2025}
}

@misc{intelligence2025pi05visionlanguageactionmodelopenworld,
      title={$\pi_{0.5}$: a Vision-Language-Action Model with Open-World Generalization}, 
      author={Physical Intelligence and Kevin Black and Noah Brown and James Darpinian and Karan Dhabalia and Danny Driess and Adnan Esmail and Michael Equi and Chelsea Finn and others},
      year={2025},
      eprint={2504.16054},
      archivePrefix={arXiv},
      primaryClass={cs.LG},
      url={https://arxiv.org/abs/2504.16054}, 
}

@article{chen2025janus,
  title={Janus-pro: Unified multimodal understanding and generation with data and model scaling},
  author={Chen, Xiaokang and Wu, Zhiyu and Liu, Xingchao and Pan, Zizheng and Liu, Wen and Xie, Zhenda and Yu, Xingkai and Ruan, Chong},
  journal={arXiv preprint arXiv:2501.17811},
  year={2025}
}

@inproceedings{karamcheti2024prismatic,
  title={Prismatic vlms: Investigating the design space of visually-conditioned language models},
  author={Karamcheti, Siddharth and Nair, Suraj and Balakrishna, Ashwin and Liang, Percy and Kollar, Thomas and Sadigh, Dorsa},
  booktitle={Forty-first International Conference on Machine Learning},
  year={2024}
}

@article{padalkar2023guided,
  title={A guided reinforcement learning approach using shared control templates for learning manipulation skills in the real world},
  author={Padalkar, Abhishek and Quere, Gabriel and Raffin, Antonin and Silv{\'e}rio, Jo{\~a}o and Stulp, Freek},
  year={2023}
}

@misc{oh2023pr2utokyodatasets,
  author={Jihoon Oh and Naoaki Kanazawa and Kento Kawaharazuka},
  title={X-Embodiment U-Tokyo PR2 Datasets},
  year={2023},
  url={https://github.com/ojh6404/rlds_dataset_builder},
}

@misc{matsushima2023weblab,
  title={Weblab xArm Dataset},
  author={Tatsuya Matsushima and Hiroki Furuta and Yusuke Iwasawa and Yutaka Matsuo},
  year={2023},
}

@article{lin2025onetwovla,
  title={OneTwoVLA: A Unified Vision-Language-Action Model with Adaptive Reasoning},
  author={Lin, Fanqi and Nai, Ruiqian and Hu, Yingdong and You, Jiacheng and Zhao, Junming and Gao, Yang},
  journal={arXiv preprint arXiv:2505.11917},
  year={2025}
}

@article{liu2025mla,
  title={MLA: A Multisensory Language-Action Model for Multimodal Understanding and Forecasting in Robotic Manipulation},
  author={Liu, Zhuoyang and Liu, Jiaming and Xu, Jiadong and Han, Nuowei and Gu, Chenyang and Chen, Hao and Zhou, Kaichen and Zhang, Renrui and Hsieh, Kai Chin and Wu, Kun and others},
  journal={arXiv preprint arXiv:2509.26642},
  year={2025}
}

@article{huang2025thinkact,
  title={Thinkact: Vision-language-action reasoning via reinforced visual latent planning},
  author={Huang, Chi-Pin and Wu, Yueh-Hua and Chen, Min-Hung and Wang, Yu-Chiang Frank and Yang, Fu-En},
  journal={arXiv preprint arXiv:2507.16815},
  year={2025}
}

@article{gu2025manualvla,
  title={ManualVLA: A Unified VLA Model for Chain-of-Thought Manual Generation and Robotic Manipulation},
  author={Gu, Chenyang and Liu, Jiaming and Chen, Hao and Huang, Runzhong and Wuwu, Qingpo and Liu, Zhuoyang and Li, Xiaoqi and Li, Ying and Zhang, Renrui and Jia, Peng and others},
  journal={arXiv preprint arXiv:2512.02013},
  year={2025}
}

@misc{zawalski2025roboticcontrolembodiedchainofthought,
      title={Robotic Control via Embodied Chain-of-Thought Reasoning}, 
      author={Michał Zawalski and William Chen and Karl Pertsch and Oier Mees and Chelsea Finn and Sergey Levine},
      year={2025},
      eprint={2407.08693},
      archivePrefix={arXiv},
      primaryClass={cs.RO},
      url={https://arxiv.org/abs/2407.08693}, 
}

@article{kareeremergence,
  title={Emergence of Human to Robot Transfer in Vision-Language-Action Models},
  author={Kareer, Simar and Pertsch, Karl and Darpinian, James and Hoffman, Judy and Xu, Danfei and Levine, Sergey and Finn, Chelsea and Nair, Suraj}
}

@article{zheng2026egoscale,
  title={Egoscale: Scaling dexterous manipulation with diverse egocentric human data},
  author={Zheng, Ruijie and Niu, Dantong and Xie, Yuqi and Wang, Jing and Xu, Mengda and Jiang, Yunfan and Casta{\~n}eda, Fernando and Hu, Fengyuan and Tan, You Liang and Fu, Letian and others},
  journal={arXiv preprint arXiv:2602.16710},
  year={2026}
}

@inproceedings{bi2026h,
  title={H-rdt: Human manipulation enhanced bimanual robotic manipulation},
  author={Bi, Hongzhe and Wu, Lingxuan and Lin, Tianwei and Tan, Hengkai and Su, Zhizhong and Su, Hang and Zhu, Jun},
  booktitle={Proceedings of the AAAI Conference on Artificial Intelligence},
  volume={40},
  number={22},
  pages={18135--18143},
  year={2026}
}

@inproceedings{kareer2025egomimic,
  title={Egomimic: Scaling imitation learning via egocentric video},
  author={Kareer, Simar and Patel, Dhruv and Punamiya, Ryan and Mathur, Pranay and Cheng, Shuo and Wang, Chen and Hoffman, Judy and Xu, Danfei},
  booktitle={2025 IEEE International Conference on Robotics and Automation (ICRA)},
  pages={13226--13233},
  year={2025},
  organization={IEEE}
}

@inproceedings{guzey2025bridging,
  title={Bridging the human to robot dexterity gap through object-oriented rewards},
  author={Guzey, Irmak and Dai, Yinlong and Savva, Georgy and Bhirangi, Raunaq and Pinto, Lerrel},
  booktitle={2025 IEEE International Conference on Robotics and Automation (ICRA)},
  pages={3344--3351},
  year={2025},
  organization={IEEE}
}

@article{lepert2025phantom,
  title={Phantom: Training robots without robots using only human videos},
  author={Lepert, Marion and Fang, Jiaying and Bohg, Jeannette},
  journal={arXiv preprint arXiv:2503.00779},
  year={2025}
}

@inproceedings{bharadhwaj2024track2act,
  title={Track2act: Predicting point tracks from internet videos enables generalizable robot manipulation},
  author={Bharadhwaj, Homanga and Mottaghi, Roozbeh and Gupta, Abhinav and Tulsiani, Shubham},
  booktitle={European Conference on Computer Vision},
  pages={306--324},
  year={2024},
  organization={Springer}
}

@misc{bai2025qwen3vltechnicalreport,
      title={Qwen3-VL Technical Report}, 
      author={Shuai Bai and Yuxuan Cai and Ruizhe Chen and Keqin Chen and Xionghui Chen and Zesen Cheng and Lianghao Deng and Wei Ding and Chang Gao and Chunjiang Ge and Wenbin Ge and Zhifang Guo and Qidong Huang and Jie Huang and Fei Huang and Binyuan Hui and Shutong Jiang and Zhaohai Li and Mingsheng Li and Mei Li and Kaixin Li and Zicheng Lin and Junyang Lin and Xuejing Liu and Jiawei Liu and Chenglong Liu and Yang Liu and Dayiheng Liu and Shixuan Liu and Dunjie Lu and Ruilin Luo and Chenxu Lv and Rui Men and Lingchen Meng and Xuancheng Ren and Xingzhang Ren and Sibo Song and Yuchong Sun and Jun Tang and Jianhong Tu and Jianqiang Wan and Peng Wang and Pengfei Wang and Qiuyue Wang and Yuxuan Wang and Tianbao Xie and Yiheng Xu and Haiyang Xu and Jin Xu and Zhibo Yang and Mingkun Yang and Jianxin Yang and An Yang and Bowen Yu and Fei Zhang and Hang Zhang and Xi Zhang and Bo Zheng and Humen Zhong and Jingren Zhou and Fan Zhou and Jing Zhou and Yuanzhi Zhu and Ke Zhu},
      year={2025},
      eprint={2511.21631},
      archivePrefix={arXiv},
      primaryClass={cs.CV},
      url={https://arxiv.org/abs/2511.21631}, 
}

@article{beyer2024paligemma,
  title={Paligemma: A versatile 3b vlm for transfer},
  author={Beyer, Lucas and Steiner, Andreas and Pinto, Andr{\'e} Susano and Kolesnikov, Alexander and Wang, Xiao and Salz, Daniel and Neumann, Maxim and Alabdulmohsin, Ibrahim and Tschannen, Michael and Bugliarello, Emanuele and others},
  journal={arXiv preprint arXiv:2407.07726},
  year={2024}
}

@article{gao2026dreamdojo,
  title={DreamDojo: A Generalist Robot World Model from Large-Scale Human Videos},
  author={Gao, Shenyuan and Liang, William and Zheng, Kaiyuan and Malik, Ayaan and Ye, Seonghyeon and Yu, Sihyun and Tseng, Wei-Cheng and Dong, Yuzhu and Mo, Kaichun and Lin, Chen-Hsuan and others},
  journal={arXiv preprint arXiv:2602.06949},
  year={2026}
}

@article{qiu2025humanoid,
  title={Humanoid policy\~{} human policy},
  author={Qiu, Ri-Zhao and Yang, Shiqi and Cheng, Xuxin and Chawla, Chaitanya and Li, Jialong and He, Tairan and Yan, Ge and Yoon, David J and Hoque, Ryan and Paulsen, Lars and others},
  journal={arXiv preprint arXiv:2503.13441},
  year={2025}
}

@article{liu2026last,
  title={LaST $ \_ $\{$0$\}$ $: Latent Spatio-Temporal Chain-of-Thought for Robotic Vision-Language-Action Model},
  author={Liu, Zhuoyang and Liu, Jiaming and Chen, Hao and Yu, Jiale and Guo, Ziyu and Hou, Chengkai and Gu, Chenyang and Mi, Xiangju and Zhang, Renrui and Wu, Kun and others},
  journal={arXiv preprint arXiv:2601.05248},
  year={2026}
}

@article{generalist2026gen1,
author = {Generalist AI Team},
title = {GEN-1: Scaling Embodied Foundation Models to Mastery},
journal = {Generalist AI Blog},
year = {2026},
note = {https://generalistai.com/blog/apr-02-2026-GEN-1},
}

@article{intelligence2026pi,
  title={$\pi_{0.7}$: A Steerable Generalist Robotic Foundation Model with Emergent Capabilities},
  author={Intelligence, Physical and Ai, Bo and Amin, Ali and Aniceto, Raichelle and Balakrishna, Ashwin and Balke, Greg and Black, Kevin and Bokinsky, George and Cao, Shihao and Charbonnier, Thomas and others},
  journal={arXiv preprint arXiv:2604.15483},
  year={2026}
}

@misc{from_human_skill_to_robotic_mastery_2026,
  title        = {From Human Skill to Robotic Mastery},
  author       = {{PsiBot Team}},
  year         = {2026},
  month        = mar,
  day          = {31},
  howpublished = {\url{https://cypypccpy.github.io/tech-blog.github.io/}},
  note         = {Accessed: 2026-05-20}
}

@article{guo2025ctrl,
  title={Ctrl-world: A controllable generative world model for robot manipulation},
  author={Guo, Yanjiang and Shi, Lucy Xiaoyang and Chen, Jianyu and Finn, Chelsea},
  journal={arXiv preprint arXiv:2510.10125},
  year={2025}
}

@article{wang2023mimicplay,
  title={Mimicplay: Long-horizon imitation learning by watching human play},
  author={Wang, Chen and Fan, Linxi and Sun, Jiankai and Zhang, Ruohan and Fei-Fei, Li and Xu, Danfei and Zhu, Yuke and Anandkumar, Anima},
  journal={arXiv preprint arXiv:2302.12422},
  year={2023}
}

@article{park2024dexhub,
  title={Dexhub and dart: Towards internet scale robot data collection},
  author={Park, Younghyo and Bhatia, Jagdeep Singh and Ankile, Lars and Agrawal, Pulkit},
  journal={arXiv preprint arXiv:2411.02214},
  year={2024}
}

@article{duan2023ar2d2,
  title={Ar2-d2: Training a robot without a robot},
  author={Duan, Jiafei and Wang, Yi Ru and Shridhar, Mohit and Fox, Dieter and Krishna, Ranjay},
  journal={arXiv preprint arXiv:2306.13818},
  year={2023}
}

@article{tao2025dexwild,
  title={Dexwild: Dexterous human interactions for in-the-wild robot policies},
  author={Tao, Tony and Srirama, Mohan Kumar and Liu, Jason Jingzhou and Shaw, Kenneth and Pathak, Deepak},
  journal={arXiv preprint arXiv:2505.07813},
  year={2025}
}

@article{yang2025egovla,
  title={Egovla: Learning vision-language-action models from egocentric human videos},
  author={Yang, Ruihan and Yu, Qinxi and Wu, Yecheng and Yan, Rui and Li, Borui and Cheng, An-Chieh and Zou, Xueyan and Fang, Yunhao and Cheng, Xuxin and Qiu, Ri-Zhao and others},
  journal={arXiv preprint arXiv:2507.12440},
  year={2025}
}

@article{kim2025fine,
  title={Fine-tuning vision-language-action models: Optimizing speed and success},
  author={Kim, Moo Jin and Finn, Chelsea and Liang, Percy},
  journal={arXiv preprint arXiv:2502.19645},
  year={2025}
}

@article{cai2026internvla,
  title={InternVLA-A1: Unifying Understanding, Generation and Action for Robotic Manipulation},
  author={Cai, Junhao and Cai, Zetao and Cao, Jiafei and Chen, Yilun and He, Zeyu and Jiang, Lei and Li, Hang and Li, Hengjie and Li, Yang and Liu, Yufei and others},
  journal={arXiv preprint arXiv:2601.02456},
  year={2026}
}

@article{lyu2026lda,
  title={Lda-1b: Scaling latent dynamics action model via universal embodied data ingestion},
  author={Lyu, Jiangran and Liu, Kai and Zhang, Xuheng and Liao, Haoran and Feng, Yusen and Zhu, Wenxuan and Shen, Tingrui and Chen, Jiayi and Zhang, Jiazhao and Dong, Yifei and others},
  journal={arXiv preprint arXiv:2602.12215},
  year={2026}
}

@inproceedings{
su2026freqpolicy,
title={FreqPolicy: Efficient Flow-based Visuomotor Policy via Frequency Consistency},
author={Yifei Su and Ning Liu and Dong Chen and Zhen Zhao and Kun Wu and Meng Li and Zhiyuan Xu and Zhengping Che and Jian Tang},
booktitle={The Thirty-ninth Annual Conference on Neural Information Processing Systems},
year={2025},
}

@inproceedings{banerjee2025hot3d,
  title={Hot3d: Hand and object tracking in 3d from egocentric multi-view videos},
  author={Banerjee, Prithviraj and Shkodrani, Sindi and Moulon, Pierre and Hampali, Shreyas and Han, Shangchen and Zhang, Fan and Zhang, Linguang and Fountain, Jade and Miller, Edward and Basol, Selen and others},
  booktitle={Proceedings of the IEEE/CVF Conference on Computer Vision and Pattern Recognition},
  pages={7061--7071},
  year={2025}
}

@inproceedings{grauman2024ego,
  title={Ego-exo4d: Understanding skilled human activity from first-and third-person perspectives},
  author={Grauman, Kristen and Westbury, Andrew and Torresani, Lorenzo and Kitani, Kris and Malik, Jitendra and Afouras, Triantafyllos and Ashutosh, Kumar and Baiyya, Vijay and Bansal, Siddhant and Boote, Bikram and others},
  booktitle={Proceedings of the IEEE/CVF Conference on Computer Vision and Pattern Recognition},
  pages={19383--19400},
  year={2024}
}

@article{xu2026compliant,
  title={Compliant residual dagger: Improving real-world contact-rich manipulation with human corrections},
  author={Xu, Xiaomeng and Hou, Yifan and Liu, Zeyi and Song, Shuran},
  journal={Advances in Neural Information Processing Systems},
  volume={38},
  pages={139559--139581},
  year={2026}
}

@article{kim2026cosmos,
  title={Cosmos Policy: Fine-Tuning Video Models for Visuomotor Control and Planning},
  author={Kim, Moo Jin and Gao, Yihuai and Lin, Tsung-Yi and Lin, Yen-Chen and Ge, Yunhao and Lam, Grace and Liang, Percy and Song, Shuran and Liu, Ming-Yu and Finn, Chelsea and Gu, Jinwei},
  journal={arXiv preprint arXiv:2601.16163},
  year={2026}
}

@inproceedings{zhang2025hawor,
  title={Hawor: World-space hand motion reconstruction from egocentric videos},
  author={Zhang, Jinglei and Deng, Jiankang and Ma, Chao and Potamias, Rolandos Alexandros},
  booktitle={Proceedings of the Computer Vision and Pattern Recognition Conference},
  pages={1805--1815},
  year={2025}
}

@article{lei2026mechanistic,
  title={A Mechanistic Analysis of Sim-and-Real Co-Training in Generative Robot Policies},
  author={Lei, Yu and Liu, Minghuan and Maddukuri, Abhiram and Jiang, Zhenyu and Zhu, Yuke},
  journal={arXiv preprint arXiv:2604.13645},
  year={2026}
}

@article{yang2025qwen3,
  title={Qwen3 technical report},
  author={Yang, An and Li, Anfeng and Yang, Baosong and Zhang, Beichen and Hui, Binyuan and Zheng, Bo and Yu, Bowen and Gao, Chang and Huang, Chengen and Lv, Chenxu and others},
  journal={arXiv preprint arXiv:2505.09388},
  year={2025}
}
